\documentclass[runningheads]{llncs}
\usepackage{graphicx}

\usepackage{tikz}
\usepackage{comment}
\usepackage{amsmath,amssymb} 
\usepackage{color}

\usepackage{marvosym} 

\usepackage{multicol}
\usepackage{multirow}
\usepackage{graphics}
\usepackage{booktabs} 

\def\blue#1{\textbf{\color{blue} #1}} 
\def\red#1{\textbf{\color{red}\underline{#1}}} 

\renewcommand{\lastandname}{\unskip,}

\usepackage[accsupp]{axessibility}  

\begin{document}
\pagestyle{headings}
\mainmatter

\title{DFTR: Depth-supervised Fusion Transformer for\\ Salient Object Detection} 


\titlerunning{DFTR: Depth-supervised Fusion Transformer}
\author{Heqin Zhu\inst{1,2} \and Xu Sun\inst{2}\;\textsuperscript{\Letter} \and Yuexiang Li\inst{2} \and Kai Ma\inst{2} \and S. Kevin Zhou\inst{1,3}\;\textsuperscript{\Letter} \and \\ Yefeng Zheng\inst{2} }
	    
\authorrunning{Zhu et al.}
%
\institute{Key Lab of Intelligent Information Processing of Chinese Academy of Sciences,\\
		Institute of Computing Technology, CAS, Beijing 100190, China \and
	    Tencent Jarvis Lab, Shenzhen, China\and
	    School of Biomedical Engineering \& Suzhou Institute for Advanced Research,\\
		Center for Medical Imaging, Robotics, and Analytic Computing \& LEarning (MIRACLE),\\
		University of Science and Technology of China, Suzhou 215123, China\\
\email{ericxsun@tencent.com, s.kevin.zhou@gmail.com}}

\maketitle

\begin{abstract}
 Automated salient object detection (SOD) plays an increasingly crucial role in many computer vision applications. By reformulating the depth information as supervision rather than as input, depth-supervised convolutional neural networks (CNN) have achieved promising results on both RGB and RGB-D SOD scenarios with the merits of no requirements for extra depth networks and depth inputs in the inference stage. This paper, for the first time, seeks to expand the applicability of depth supervision to the Transformer architecture. Specifically, we develop a Depth-supervised Fusion TRansformer (DFTR), to further improve the accuracy of both RGB and RGB-D SOD. The proposed DFTR involves three primary features: 1) DFTR, to the best of our knowledge, is the first pure Transformer-based model for depth-supervised SOD; 2) A multi-scale feature aggregation (MFA) module is proposed to fully exploit the multi-scale features encoded by the Swin Transformer in a coarse-to-fine manner; 3) To enable bidirectioal information flow across different streams of features, a novel multi-stage feature fusion (MFF) module is further integrated into our DFTR with the emphasis on salient regions at different network learning stages. We extensively evaluate the proposed DFTR on ten benchmarking datasets. Experimental results show that our DFTR consistently outperforms the existing state-of-the-art methods for both RGB and RGB-D SOD tasks. The code and model will be made publicly available.
\end{abstract}

\section{Introduction} \label{sec:intro}
\begin{figure}[t]
    \centering
    \includegraphics[width=0.7\linewidth]{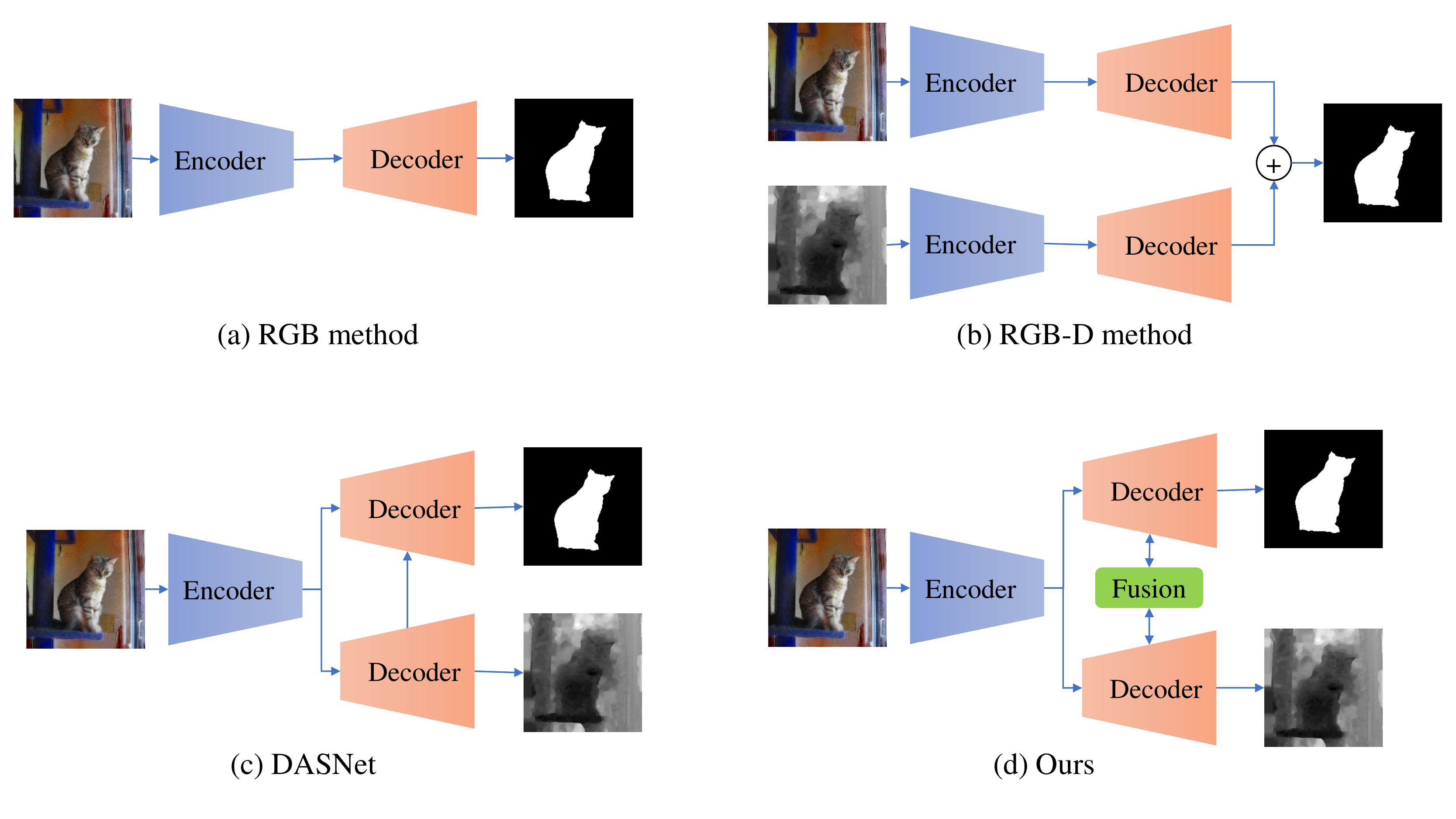}
    \caption{Different kinds of SOD frameworks. (a) Typical RGB-based: with RGB as the only input. (b) Typical RGB-D-based: with both RGB and depth as inputs. (c) DASNet~\cite{dasnet}: with RGB as input and depth as supervision, decoding information only feeds from the depth map recovery branch to the salient object segmentation branch. (d) Ours: with RGB as input and depth as supervision, a multi-stage feature fusion module is further developed to enable bidirectional information flow from different streams of features.
    }
    \label{fig_overview}
\end{figure}

Salient object detection (SOD), which aims to detect and segment the most noticeable objects in a scene, is a fundamental task for various computer vision applications, such as semantic segmentation~\cite{Lee_2021_CVPR}, image translation~\cite{Jiang_2021_CVPR}  and visual tracking~\cite{lee2021sspnet}. At the very beginning of this research line, only the RGB images were taken as input (Fig.~\ref{fig_overview} (a)). For example, Zhao \emph{et al.} \cite{Zhao_2019_CVPR} proposed a pyramid feature attention network that generates saliency maps. Although the RGB-based approaches are simple and easy-to-implement, they often fail to tackle images with complex backgrounds.

To address the problem, researchers attempted to integrate the depth information into SOD frameworks as complementary guidance for object detection. As shown in Fig.~\ref{fig_overview} (b), most RGB-D frameworks deployed two streams to extract useful features from RGB and depth images, respectively, and then fused the features for the plausible saliency results. For example, Han \emph{et al.} \cite{8091125} proposed a two-stream network to extract features from RGB and depth images, respectively, and then fuse them with a combination layer. Zhao \emph{et al.} \cite{Zhao_2019a_CVPR} implemented a feature-enhancement module and fluid pyramid integration module for the better fusion of RGB and depth features. Owing to the auxiliary depth information, the accuracy of SOD task gains a significant increase. However, all these methods require extra depth networks and depth inputs during inference, and thus reveal two main drawbacks: 1) the additional depth branch increases the computational cost and 2) the paired depth maps are usually unavailable in real applications.

To deal with the aforementioned challenges, a new research line, which only takes the depth maps as supervision for SOD at training time, has drawn increasing attentions in the community. For example, Piao {\it et al.}~\cite{piao2020} proposed a depth distiller to transfer the depth knowledge from the depth stream to the RGB stream, which results in a lightweight architecture free of the depth stream at test time. Ji {\it et al.}~\cite{Wei_2020_ECCV}, instead, developed a novel collaborative learning framework (CoNet) to bypass the use of additional depth network and depth input during testing. Zhao \emph{et al.}~\cite{dasnet} further formulated a multi-task network to simultaneously perform saliency detection and depth estimation. The auxiliary information learned from the depth map recovery task is adopted to regularize the features for saliency segmentation and thereby boosts the SOD performance. However, all previous depth-supervised methods are based on vanilla convolutional neural networks (CNN) and can fail to exploit global long-range dependencies. From another aspect, motivated by the global cues modeling capabilities of vision Transformers (ViT)~\cite{vit}, a recent study introduced a unified model termed Visual Saliency Transformer (VST)~\cite{vst} for both RGB and RGB-D tasks. Although VST has shown to surpass all CNN-based depth-supervised models, its reliance on an auxiliary input to incorporate the additional depth information limits its applications. Therefore, we raise an open and valuable question: \emph{is it possible to reclaim the superiority of depth-supervised paradigm through a novel design of pure Transformer?}

To answer this question, we propose a \textbf{D}epth-supervised \textbf{F}usion \textbf{TR}ansformer (\textbf{DFTR}) for both RGB and RGB-D SOD, which, to the best of our knowledge, is the first attempt to expand the applicability of depth supervision to the pure Transformer architecture. Our DFTR has an encoder and a two-stream decoder to jointly perform salient object detection and depth map prediction, as shown in Fig.~\ref{fig_overview}.
In particular, the proposed DFTR adopts Swin Transformer~\cite{swin} as the backbone of encoder to effectively extract discriminative features, and its decoder consists of two novel modules (\emph{i.e.}, multi-scale feature aggregation (MFA) and multi-stage feature fusion (MFF)) for the dense high-resolution prediction. The former module aggregates features from the adjacent scales, while the latter one improves the flexibility of information flow in a hierarchical manner. Our main contributions can be summarized as follows:
\begin{itemize}
    \item For the first time, we propose a depth-supervised SOD network built upon a pure Transformer architecture. The proposed DFTR adopts the depth supervision learning strategy; hence, only the RGB image is required at inference time for both RGB and RGB-D SOD tasks.
    \item Two simple-yet-effective Transformer-based modules, {\it i.e.}, MFA and MFF, are developed to instinctively fuse features extracted from adjacent scales and different decoder streams, respectively. 
    \item The proposed DFTR framework is evaluated on multiple publicly available SOD benchmarking datasets. The experimental results validate the effectiveness of our DFTR --- the superiority of depth-supervised framework for SOD is reclaimed.
\end{itemize}

\section{Related Work}\label{sec:relat}
\subsection{Transformer}
Due to the self-attention mechanism, Transformer can effectively model long-range dependencies and extract global context features. In recent years, Transformer has shown the superiority to convolutional neural network (CNN) in various computer vision tasks, \emph{e.g.}, image classification~\cite{vit,Chen_2021_ICCV,zhang2021token}, object detection~\cite{detr,sheng2021improving}, and image segmentation~\cite{setr,wang2021end,strudel2021segmenter}. A variety of pure Transformer backbone networks without convolution operations have been proposed. For example, 
T2T-ViT~\cite{t2t-vit} incorporates a layer-wise Tokens-to-Token (T2T) transformation with a deep-narrow efficient backbone to simultaneously reduce token length and extract rich local features. The Swin Transformer~\cite{swin} decreases the time complexity of self-attention computation from $\mathcal{O}(N^2)$ to $\mathcal{O}(N)$ by limiting the areas for self-attention computation (\emph{i.e.}, a local window). Such a local window will be shifted across the whole image to capture the global context.

\subsection{Salient Object Detection}

\subsubsection{CNN-based SOD Methods.}
For RGB SOD and RGB-D SOD tasks, various CNN-based methods~\cite{MINet-R,LDF-R,CSF-R2,gatenet,DHNet,zhaoeccv20,dasnet,bbsnet,fan2019rethinking,jldcf,dfmnet,btsnet,li2020nui} have been proposed in recent years, which are superior to traditional methods~\cite{feng2016local,cong2016saliency,guo2016salient,song2017depth,zhu2017innovative,cong2019going} based on hand-crafted features. Most CNN-based RGB SOD methods~\cite{MINet-R,LDF-R,CSF-R2,dasnet,bbsnet,btsnet,li2020nui} adopt the encoder-decoder architecture. Commonly, the encoder uses a pre-trained network (\emph{e.g.}, ResNet~\cite{resnet} and VGG~\cite{vgg}) as backbone, while the decoder is an elaborately designed network. For examples, Zhang \emph{et al.}~\cite{zhang2020uc} proposed the first uncertainty model, which has a generative architecture to learn from data labeling process. Fu \emph{et al.}~\cite{jldcf} developed a Siamese network to jointly learn saliency map and depth map. A novel densely cooperative fusion module was proposed to extract complementary features. Tang \emph{et al.}~\cite{DHNet} disentangled the SOD task into a low-resolution saliency classification task and a high-resolution refinement regression task. Furthermore, several studies~\cite{gatenet,bbsnet,dutlf,li2020asif} integrated the spatial and channel attention mechanism to the decoder for SOD performance improvement. Most existing CNN-based RGB-D SOD methods mainly focus on the fusion (\emph{e.g.}, summation, multiplication or concatenation) of RGB and depth features~\cite{chen2019three,jldcf,chen2018progressively} or the utilization of depth map~\cite{dcf,dutlf,Zhao_2019a_CVPR,li2020icnet,cong2019going}. 

\subsubsection{Transformer-based SOD Methods.}
Until now, there are only two Transformer-based SOD methods, termed visual saliency Transformer (VST)~\cite{vst} and TriTransNet~\cite{tritransnet}. The VST~\cite{vst} proposed by Liu \emph{et al.} was the first pure Transformer method for SOD task, which adopted T2T-ViT~\cite{t2t-vit} as backbone network. Concretely, the VST leveraged multi-level token fusion and adopted a new token upsampling method to yield high-resolution detection results. 
TriTransNet~\cite{tritransnet} had a triplet Transformer embedding module to capture long-range dependencies across layers. In particular, the encoders of triplet Transformer module share weights for multi-level feature enhancement, while the three-stream decoder is individually initialized for multi-modal fusion.
Both Transformer-based methods achieved satisfactory SOD performances, \emph{i.e.}, outperforming the state-of-the-art CNN-based approaches, which validates the effectiveness of Transformer structure.

However, all current Transformer-based methods take the additional depth images as an extra input stream during testing, which severely limits their practical applications as the paired depth maps are usually exceedingly noisy or even unavailable in real-world setups. In contrast, our DFTR, for the first time, adopts a pure Transformer-based depth-supervised learning framework for both RGB and RGB-D SOD, being free of the depth stream at test time and meanwhile maximizing the performance. In particular, two novel Transformer-based modules, namely, MFA and MFF, are further proposed to effectively utilize the hierarchical features extracted from a Swin Transformer backbone.


\section{Methodology}
\label{sec:method}
\begin{figure*}[t]
    \centering
    \includegraphics[width=0.9\linewidth]{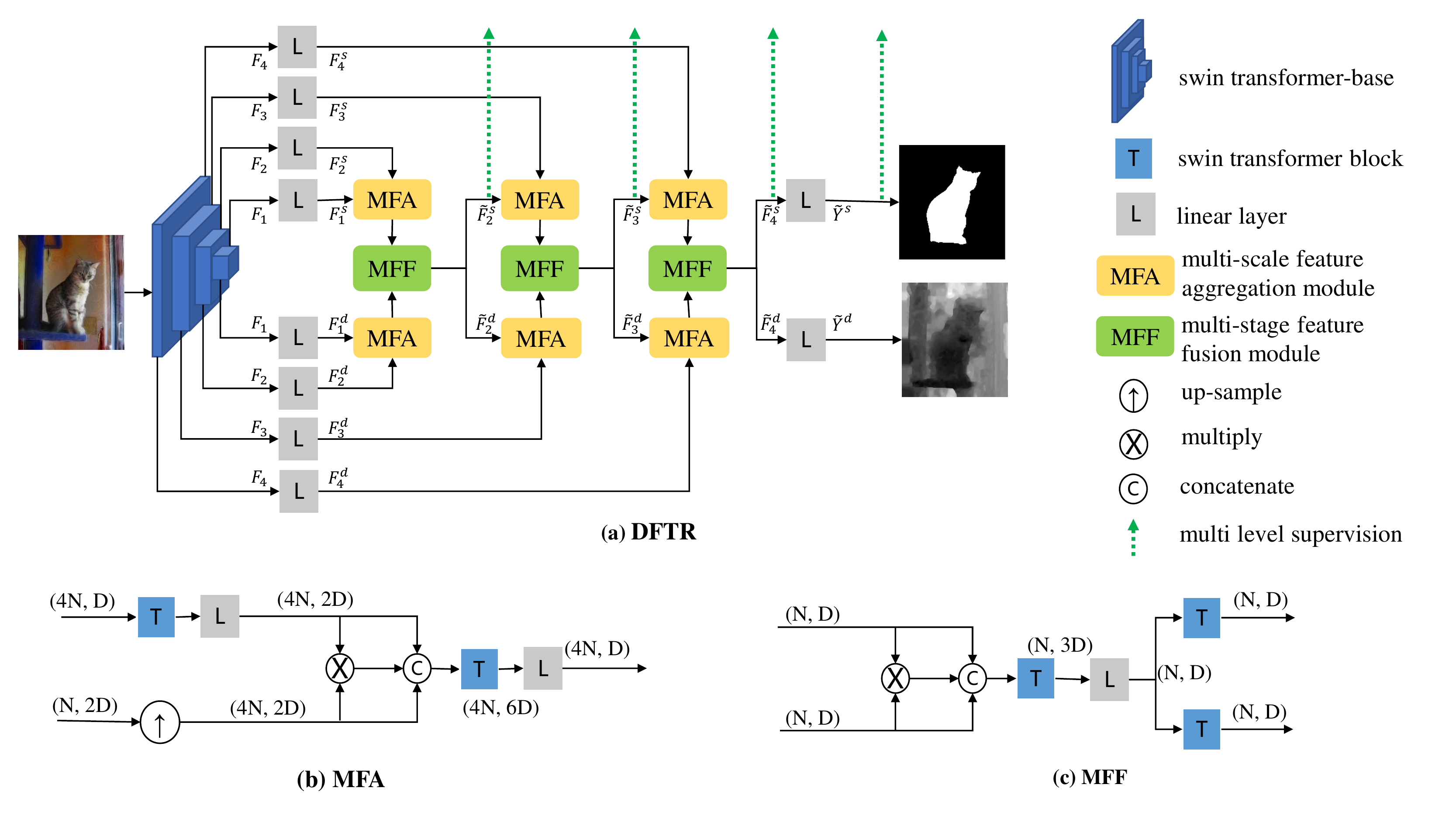}
    \caption{(a) Overview of the network structure of DFTR which consists of a Swin Transformer encoder backbone and two hierarchical decoding branches for saliency detection and depth estimation, respectively. (b) MFA: Multi-scale feature aggregation module. (c) MFF: Multi-stage feature fusion module.}
    \label{fig_network}
\end{figure*}

We present a novel pure Transformer-based neural network for both RGB and RGB-D SOD tasks, named as \textbf{D}epth-supervised \textbf{F}usion \textbf{TR}ansformer (\textbf{DFTR}). As Fig.~\ref{fig_network} shows, our proposed DFTR model takes an encoder-decoder architecture. The encoder is a Transformer network based on Swin Transformer~\cite{swin} for feature extraction, as described in Section~\ref{sec:encoder}. The decoder has a dual-steam structure for outputting saliency map and depth map, consisting of a multi-scale feature aggregation (MFA) module (Section~\ref{sec:MFA}) and a multi-stage feature fusion (MFF) module (Section~\ref{sec:MFF}). The MFA module aggregates multi-scale features extracted from encoder from high-level to low-level while the MFF module enables bidirectional information flow for joint feature learning yet with the emphasis on saliency detection at different decoding stages. The overall architecture is demonstrated in Section~\ref{sec:overall} and the learning objective is in Section~\ref{sec:loss}.

\subsection{Transformer Encoder}
\label{sec:encoder}


\subsubsection{Swin Transformer Block.}

Swin Transformer~\cite{swin} mainly consists of Transformer blocks~\cite{Transformer} in which the standard multi-head self-attention (MSA) module is replaced by a shifted window-based module. In detail, a Swin Transformer block is composed of a shifted window multi-head self-attention module (SW-MSA), followed by a two-layer multi-layer perceptron (MLP) with GELU activation. A LayerNorm (LN) layer is adopted before each SW-MSA and each MLP, and a residual shortcut is adopted after each module. In this paper, we also adopt Swin Transformer block to build the MFA and MFF modules, which are described in Section~\ref{sec:MFA} and Section~\ref{sec:MFF}, respectively.

Different from standard Transformer architecture which conducts global self-attention, Swin Transformer only performs self-attention within non-overlapping local windows for efficient modeling and rapid computation. To enable cross-window connections and enhance long-range dependencies, a shifted window partitioning approach is further introduced to shift neighboring non-overlapping window partition between consecutive Swin Transformer blocks,

\subsubsection{Hierarchical Feature Extraction.}

Swin Transformer~\cite{swin} produces four scales of hierarchical feature maps at four stages by starting from small-sized image patches and gradually merging neighboring patches in deeper layers.
First, the input RGB image is partitioned into non-overlapping patches of size $4\times 4$, resulting in $4\times 4 \times 3 = 48$ feature dimensions for each patch. 
A linear embedding layer is then employed to project this raw-valued feature into an arbitrary dimension (denoted as $C$), forming patch tokens in a shape of $(\frac{H}{4}\times \frac{W}{4}, C)$,  where $H$ and $W$ denote the height and width of the image, respectively. These patch tokens are then fed into several consecutive Swin Transformer blocks, with the number of tokens remaining as $\frac{H}{4} \times \frac{W}{4}$. The whole procedure mentioned above is referred to as ``Stage 1''.

To generate a hierarchical representation in the successive stage, patches are merged by concatenating each group of $2\times 2$ neighboring patches and then converted to high-dimension patches by a linear layer. Let $N$ denote the number of input tokens and $D$ denote the input dimension. The output shape of the patch merging layers becomes $(\frac{N}{4}, 2D)$. Similar to Stage 1, a sequence of Swin Transformer blocks is applied to the merged patches at each following stage, while keeping the number of tokens unchanged.  For Swin Transformer base model, the numbers of Swin Transformer blocks at each stage is 2, 2, 18, and 2, respectively. Finally, four levels of hierarchical feature maps, denoted as $F_4$, $F_3$, $F_2$ and $F_1$, are generated from the Swin Transformer backbone with shapes being $(\frac{H}{4}\times \frac{W}{4}, C)$, $(\frac{H}{8}\times \frac{W}{8}, 2C)$, $(\frac{H}{16}\times \frac{W}{16}, 4C)$, and $(\frac{H}{32}\times \frac{W}{32}, 8C)$, respectively.


\subsection{Multi-scale Feature Aggregation Module}
\label{sec:MFA}

The Swin Transformer encoder produces hierarchical feature maps of different spatial resolutions, but introduces large semantic gaps caused by different learning stages. The high-resolution maps contain low-level features that are very useful for accurate positioning but harm their representational power for salient object detection. In contrast, low-resolution maps contain high-level features that are semantically strong but easily blur the boundary of the salient object. In order to make full use of the low-level and high-level features for accurate dense prediction, we design a multi-scale feature aggregation (MFA) module that gradually aggregates features of different scales, with a similar form like FCN~\cite{fcn} and U-Net~\cite{unet}. 

Figure~\ref{fig_network} (b) shows the building blocks that construct our MFA module. Taking a coarse-resolution feature map and a high-resolution feature map as inputs, MFA module seeks to expand them into the same shape at first. Specifically, the spatial dimension of the coarse-resolution feature maps $\tilde{F}^{\{d,s\}}_i$ is upsampled by a factor of 2 using bilinear interpolation to generate feature maps $\tilde{F}^{\{d,s\}}_{i_{\text{mid}}}$ in shape of $(4N, 2D)$, while the channel dimension of the high-resolution feature maps $F^{\{d,s\}}_{i+1}$ is enlarged by a factor of 2 through a Swin Transformer block and a linear layer to obtain feature maps $F^{\{d,s\}}_{{i+1}_{\text{mid}}}$ in shape of $(4N, 2D)$.
Inspired by DASNet~\cite{dasnet}, we then multiply $\tilde{F}^{\{d,s\}}_{i_{\text{mid}}}$ and $F^{\{d,s\}}_{{i+1}_{\text{mid}}}$ element-wisely to form a new feature map $\tilde{F}^{\{d,s\}}_{{i+1}_{\text{mul}}}$ to enhance common pixels and alleviate ambiguous pixels. Next, we concatenate $\tilde{F}^{\{d,s\}}_{i_{\text{mid}}}$, $F^{\{d,s\}}_{{i+1}_{\text{mid}}}$ and $\tilde{F}^{\{d,s\}}_{{i+1}_{\text{mul}}}$ channel-wisely to get the aggregated feature map  $\tilde{F}^{\{d,s\}}_{{i+1}_{\text{cat}}}$ in shape of $(4N, 6D)$. Finally,  we use a Swin Transformer block to aggregate concatenated features $\tilde{F}^{\{d,s\}}_{{i+1}_{\text{cat}}}$ and a linear layer to perform channel dimension reduction to obtain the output feature maps $Z^{\{d,s\}}_{i+1}$. The above procedure is summarized as:
\begin{equation}
\begin{split}
    \tilde{F}^{\{d,s\}}_{{i+1}_{\text{cat}}}& = U( \tilde{F}^{\{d,s\}}_i) \copyright L_{\theta _2}(T_{\theta _1}(F^{\{d,s\}}_{i+1}) \copyright (U(\tilde{F}^{\{d,s\}}_i) \otimes L_{\theta _2}(T_{\theta _1}(F^{\{d,s\}}_{i+1}))),\\
    Z^{\{d,s\}}_{i+1} &= L_{\theta _4}(T_{\theta _3}(\tilde{F}^{\{d,s\}}_{{i+1}_{\text{cat}}})).
\end{split}
\end{equation}
where $\copyright$ represents concatenation; $\otimes$ represents element-wise multiplication; $U, T, L$ represent upsample, Swin Transformer block and linear layer, respectively; $i\in \{1,2,3\}$ represents different stage; $\theta _k$ with $k\in\{1,2,3,4\}$ represents learnable parameters.

\subsection{Multi-stage Feature Fusion Module}
\label{sec:MFF}

Similar to DASNet~\cite{dasnet}, we jointly perform depth estimation and saliency detection with the merits of being free of depth inputs in the inference stage. This is desirable for real-world applications as the paired depth maps are usually unavailable in real industrial setups. Motivated by the importance of flexible information flow~\cite{resnet,mergeandrun,dcf}, we further develop a novel multi-stage feature fusion (MFF) module to enable bidrectional information flow across two streams of features, which is different from DASNet where he decoded information can only be fed from the depth estimation branch to the SOD branch.

As illustrated in Fig.~\ref{fig_network} (c), the MFF module has a squeeze-and-expand structure that fuses two input feature streams, namely, $Z^s_{i} \in R^{N\times D}$  and $Z^d_{i} \in R^{N\times D}$, into $\tilde{F}^s_{i}\in R^{N\times D}$ and $\tilde{F}^d_{i}\in R^{N\times D}$. Specifically, element-wise multiplication is first performed on $Z^s_{i}$ and $Z^d_{i}$ to enhance the common features. 
Considering the spatial relevance between saliency maps and depth maps, the enhanced feature map $Z_{i_{\text{mul}}}$ is then concatenated with the source features in the channel dimension to obtain $Z_{i_{\text{cat}}}\in R^{N\times 3D}$. To further fuse the concatenated feature map, a Swin Transformer block is applied, followed by a linear layer for channel reduction. The fused feature map $Z_{i_{\text{fuse}}}$, which is dedicated to preserving the shared commonalities between two streams at different decoding stages, can be formulated as:
\begin{equation}
    Z_{i_{\text{fuse}}} =  L_{\theta _6}(T_{\theta _5}(Z^d_{i} \copyright Z^s_{i} \copyright (Z^d_{i} \otimes Z^s_{i}))).
\end{equation}

Then, we assign two independent Swin Transformer blocks to separately process the fused feature map, which results in a dual-stream output represented as:
\begin{equation}
    \tilde{F}^s_{i+1} = T_{\theta _7}(Z_{i_{\text{fuse}}}), ~~
    \tilde{F}^d_{i+1} = T_{\theta _8}(Z_{i_{\text{fuse}}}),
\end{equation}
where $\theta _k$ with $k\in\{5,6,7,8\}$ represents learnable parameters.

Moreover, we further perform the multi-level supervision (MLS) of salient regions on the upper output stream of the MFF module, as shown in Fig.~\ref{fig_network} (a). There are two reasons for this setting. First, the one-side MLS strategy prevents the MFF module from simply learning similar and swappable feature representations caused by its symmetric structure. Second, applying MLS for SOD enforces the corresponding feature extraction stream to focus more on salient regions in different network learning stages and thereby improving the final SOD performance which is consistent with the aim of this study. 

\subsection{Overview of DFTR}
\label{sec:overall}

As illustrated in Fig.~\ref{fig_network} (a), the encoded multi-scale feature maps $[F_4, F_3, F_2, F_1]$ are first passed through linear layers for channel reduction. Specifically, two groups of linear layers are employed for two decoding streams separately: the upper one for saliency detection only and the lower one for cross-task feature learning. This procedure can be defined as:
\begin{equation}
 \label{formula:in_linear}
    \begin{split}
    F_i^s = L_{\theta_i}^{s}(F_i),~~
    F_i^d = L_{\theta_i}^{d}(F_i),
    \end{split}
\end{equation}
where $s$ represents the saliency detection stream, $d$ denotes cross-task feature learning stream, and $i\in \{1,2,3,4\}$ indicates the $i$-th level. 

For each branch, three MFA modules are adopted sequentially to aggregate four levels of hierarchical features in a coarse-to-fine manner.
An MFF module is inserted at each level to fuse the outputs of MFA modules from both streams. The fused feature maps are then fed into the successive MFA modules. We summarize these operations as follows:
\begin{equation}
    \begin{split}
        \tilde{F}_1^s &= F_1^s;~~ \tilde{F}_1^d = F_1^d, \\
        Z_{i+1}^s &= {\text{MFA}^s}_{\theta_i}(\tilde{F}_i^s, F_{i+1}^s),\\
        Z_{i+1}^d &= {\text{MFA}^d}_{\theta_i}(\tilde{F}_i^d, F_{i+1}^d),\\
        \tilde{F}_{i+1}^s, \tilde{F}_{i+1}^d &= {\text{MFF}}_{\theta_i}(Z_{i+1}^s, Z_{i+1}^d), \\
    \end{split}
\end{equation}
where $i\in \{1,2,3\}$.
Finally, we use linear layers to obtain the prediction maps $\tilde{Y}^s$ and $\tilde{Y}^d$, which is formulated as:
\begin{equation}
 \label{formula:final}
    \begin{split}
    \tilde{Y}^s = L_{\theta_{\text{final}}}^{s}(\tilde{F}_4^s),~~
     \tilde{Y}^d = L_{\theta_{\text{final}}}^{d}(\tilde{F}_4^d).
    \end{split}
\end{equation}

\subsection{Learning Objective}
\label{sec:loss}
\begin{figure}[t]
    \centering
    \includegraphics[width=0.6\linewidth]{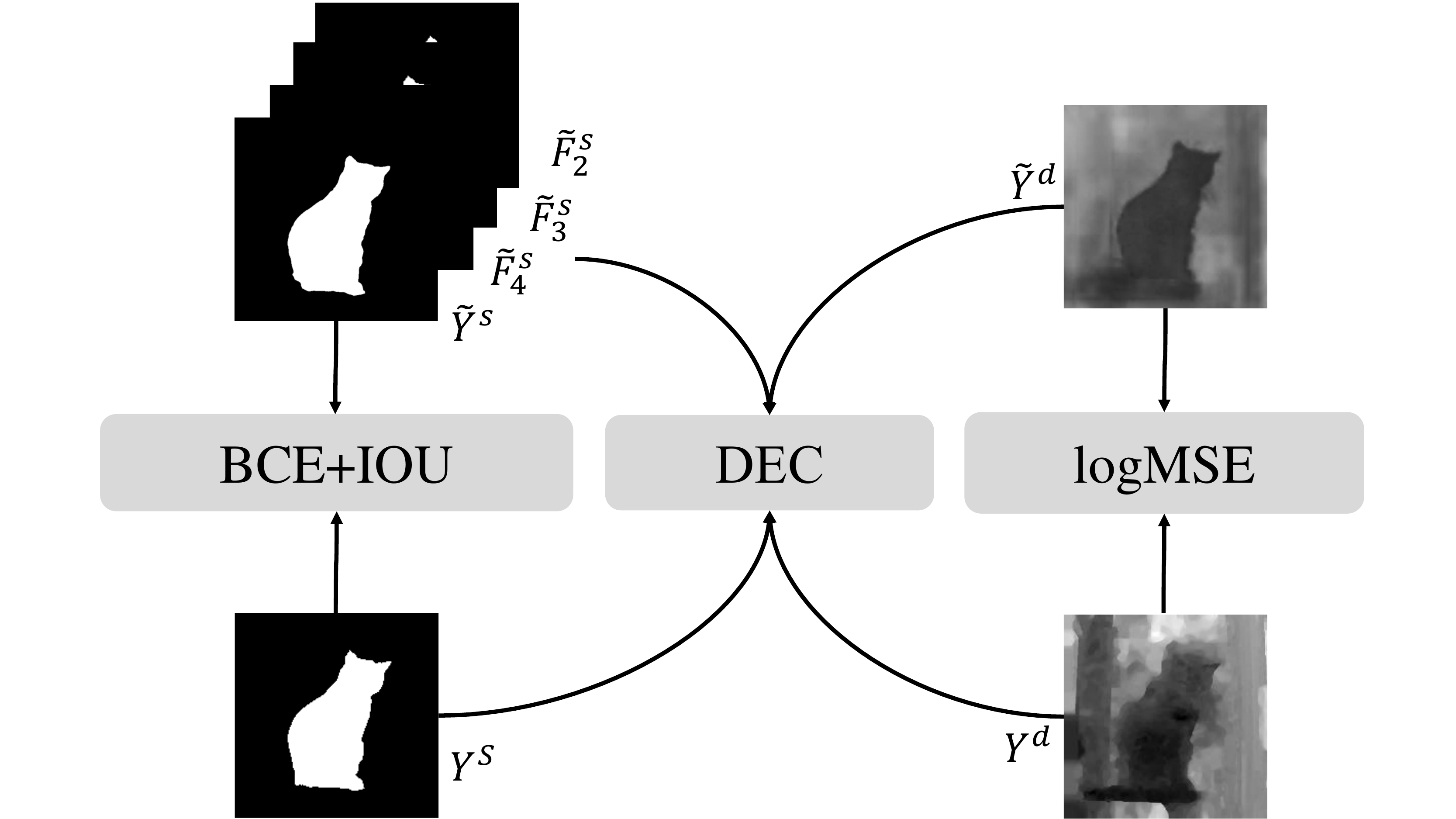} 
    \caption{An illustration of the relationships of loss functions between the predictions and ground truth labels. The DEC loss is error-weighted BCE loss from DASNet~\cite{dasnet}. $\tilde{Y}^s, \tilde{Y}^d$ are predicted saliency and depth maps, respectively, while $Y^s, Y^d$ are corresponding ground truth labels. We use additional supervision signals with saliency maps $\tilde{F}_2^s, \tilde{F}_3^s, \tilde{F}_4^s$ which are the outputs of MFF modules for multi-level supervision (MLS).}
    \label{fig:loss}
\end{figure}

Following~\cite{dasnet}, we train our network with three objectives, as illustrated in Fig.~\ref{fig:loss}. To be specific, we use binary cross entropy (BCE) and intersection over union (IOU) loss functions for SOD supervision and log mean squared error (logMSE) for depth supervision. Besides, a depth error-weighted correction (DEC)  loss~\cite{dasnet} is further adopted to mine ambiguous pixels by leveraging the depth prediction mistakes.
Three levels of intermediate saliency feature maps $\tilde{F}_2^s, \tilde{F}_3^s, \tilde{F}_4^s$ and the final prediction $Y^s$ are used for multi-level supervision (MLS). The overall loss is given as:
\begin{equation}
    \mathcal{L} = \mathcal{L}_{\text{logMSE}} + \sum_{i=1}^{4}\lambda_i (\mathcal{L}_{\text{BCE}} + \mathcal{L}_{\text{IOU}} + \mathcal{L}_{\text{DEC}}),
\end{equation}
where $\lambda_i$'s are the loss weights of all levels and empirically set to $[0.4, 0.6, 0.8, 1.0]$.

\section{Experiments}
\label{sec:exper}

\subsection{Experimental Settings}
\label{sec:setting}
\subsubsection{Datasets.} For RGB-D SOD, we adopt five publicly available datasets for performance evaluation, which are NJU2K~\cite{ju2014depth} (1,985 images), STERE~\cite{niu2012leveraging} (1,000 images), NLPR~\cite{peng2014rgbd} (1,000 images), SSD~\cite{zhu2017three} (80 images), and SIP~\cite{fan2019rethinking} (929 images). Concretely, we train SOD models with a set, consisting of 1,500 NJU2K images and 700 NLPR images. All the rest images of five benchmarking datasets are adopted for testing.

For RGB SOD, we conduct experiments on five widely-used datasets, \emph{i.e.}, DUTS~\cite{DUTS} (15,572 images), ECSSD~\cite{ECSSD} (1,000 images), DUT-OMRON~\cite{DUTOMRON} (5,168 images), PASCAL-S~\cite{PASCAL} (850 images), and HKU-IS~\cite{HKUIS} (4,447 images). Similarly, SOD models are trained with 10,553 images from the public training set of DUTS, and tested on the rest images of the five datasets. To generate the depth images of DUTS dataset for the training of our DFTR, we adopt AdaBins~\cite{adabins}, a state-of-the-art method, for depth estimation.

\subsubsection{Evaluation Metrics.}
For quantitative evaluation, we adopt four metrics to evaluate the performance of our model. Specifically, {\bf structure measure} ($S_{\alpha}$)~\cite{fan2017structure} is used to evaluate region-aware and object-aware structural similarity; {\bf maximum F-measure} ($F_{\beta}$)~\cite{achanta2009frequency} is the weighted harmonic mean of precision and recall; {\bf maximum enhanced-alignment measure} ($E_{\xi}$)~\cite{Fan2018Enhanced} jointly captures image-level statistics and local pixel matching information; {\bf mean absolute error} ($M$)~\cite{perazzi2012saliency} evaluates the pixel-wise error between predictions and the ground truth. 

\subsubsection{Implementation Details.}
\label{sec:impl}
The proposed DFTR is implemented using PyTorch. The model is trained with an NVIDIA Tesla V100 GPU (version 11).\footnote{The code will be released if the paper is accepted.} We adopt the Swin Transformer pre-trained on ImageNet~\cite{russakovsky2015imagenet} as the backbone of encoder. At the training stage, we resize each image to $352\times 352$ pixels and adopt random horizontal flipping, random cropping, and multi-scale resizing for data augmentation. The SGD optimizer is employed for network optimization. The maximum learning rate is 0.002 for backbone and 0.02 for other parts, which cyclically varies from zero to maximum and then from maximum to zero. The batch size is set to 16. Our model is observed to converge after 200 epochs of training. At testing stage, the image is first resized to $352\times 352$ pixels and then fed into our model to yield the predicted saliency map, which is finally rescaled back to the original size for SOD performance evaluation.

\subsection{Comparisons with State-of-the-art Methods}

\subsubsection{RGB-D SOD comparison.} 
\begin{figure*}[t]
    \scriptsize
    \centering
    \begin{minipage}[t]{0.1\textwidth} 
        \vspace{0pt} 
        \centering
        \includegraphics[width=1\linewidth]{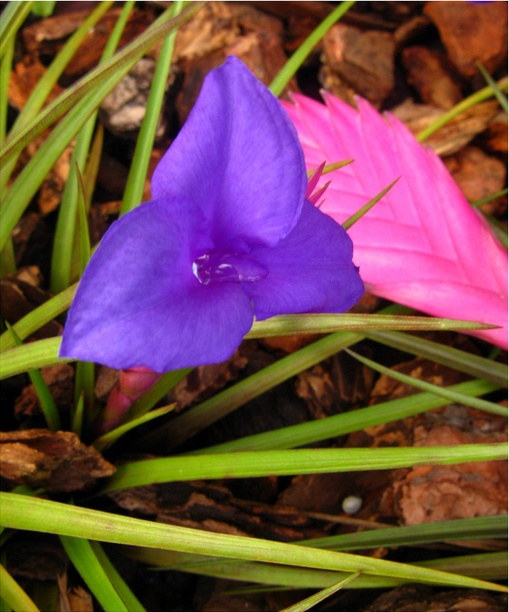}
        \includegraphics[width=1\linewidth]{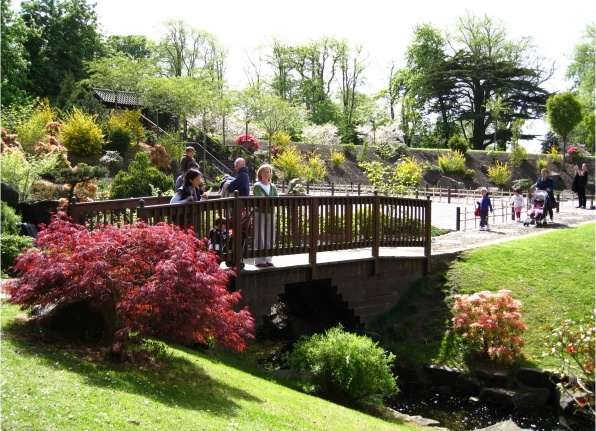}
        \includegraphics[width=1\linewidth]{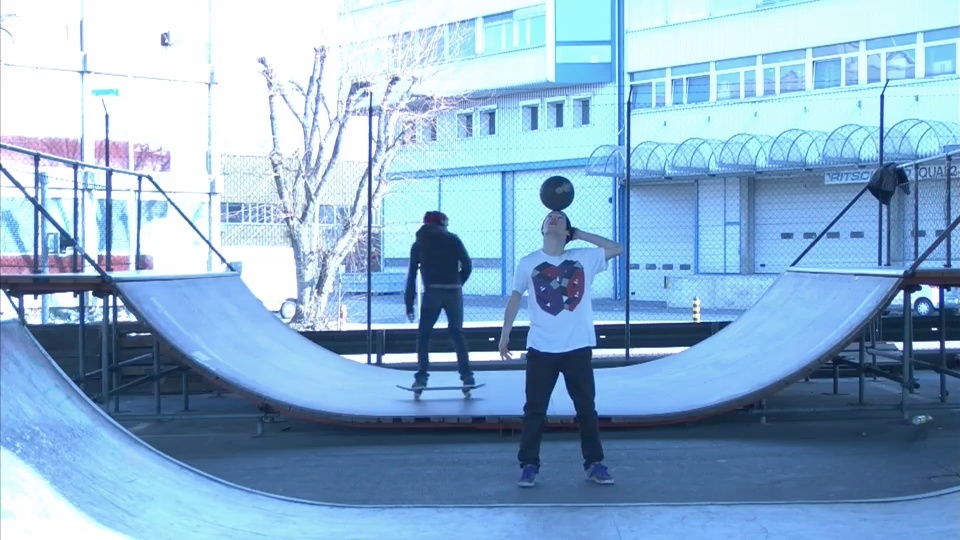}
        \includegraphics[width=1\linewidth]{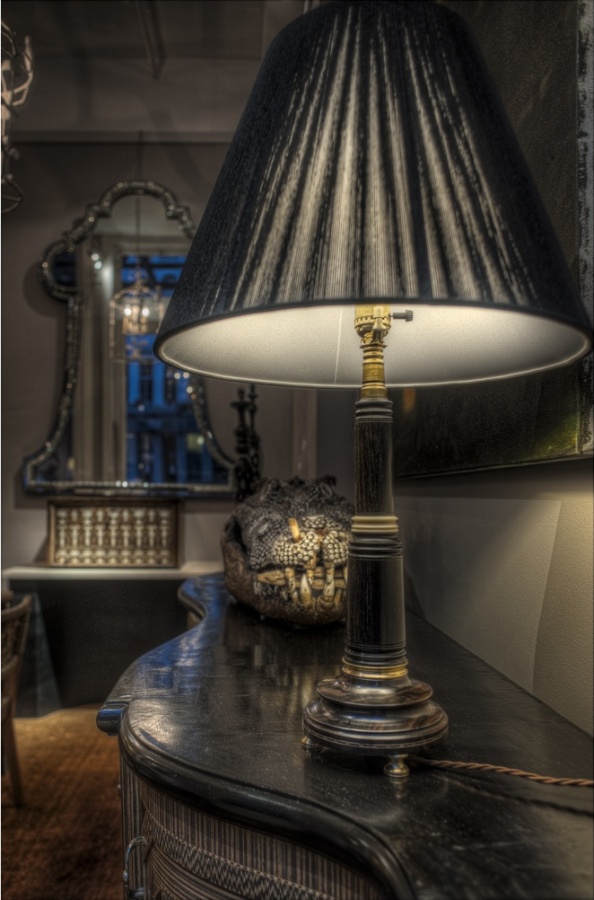}
        \includegraphics[width=1\linewidth]{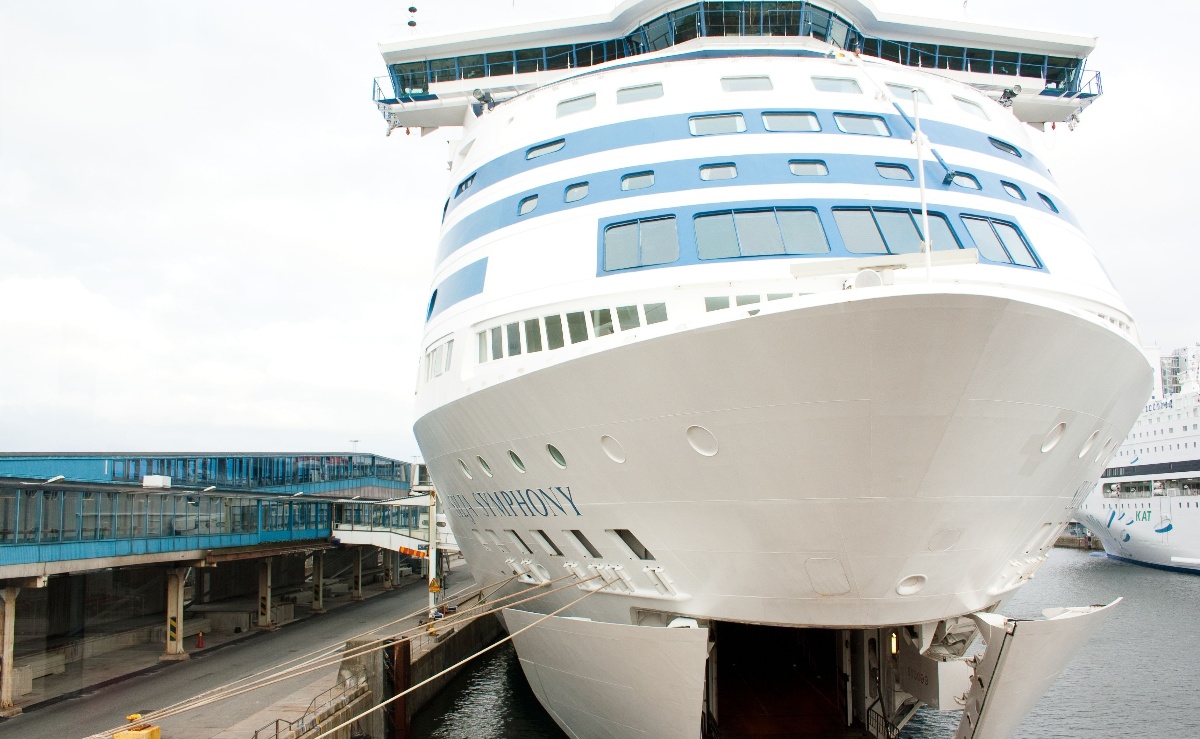}
        RGB
    \end{minipage}
    \begin{minipage}[t]{0.1\textwidth}
    \vspace{0pt}
        \centering
        \includegraphics[width=1\linewidth]{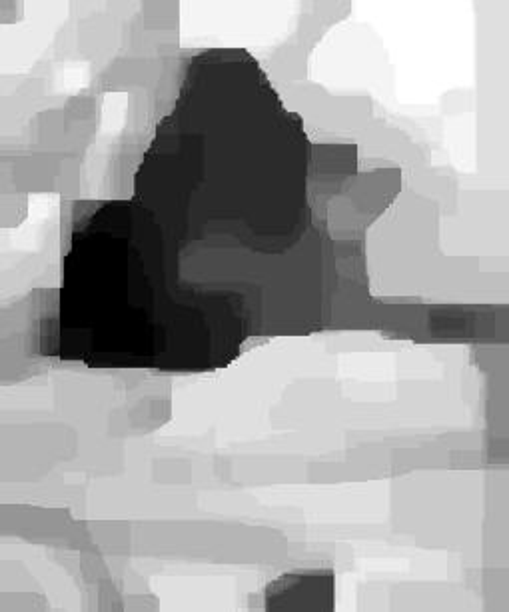}
        \includegraphics[width=1\linewidth]{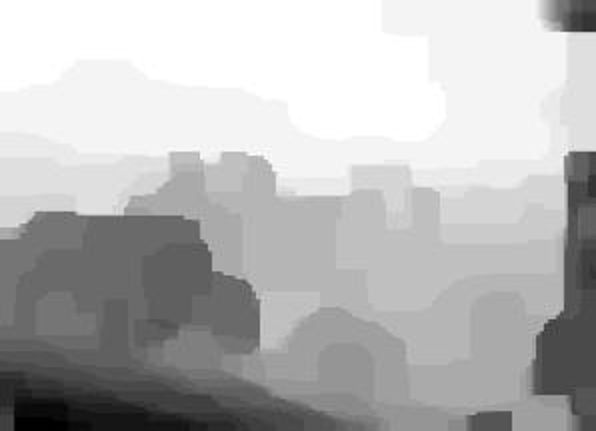}
        \includegraphics[width=1\linewidth]{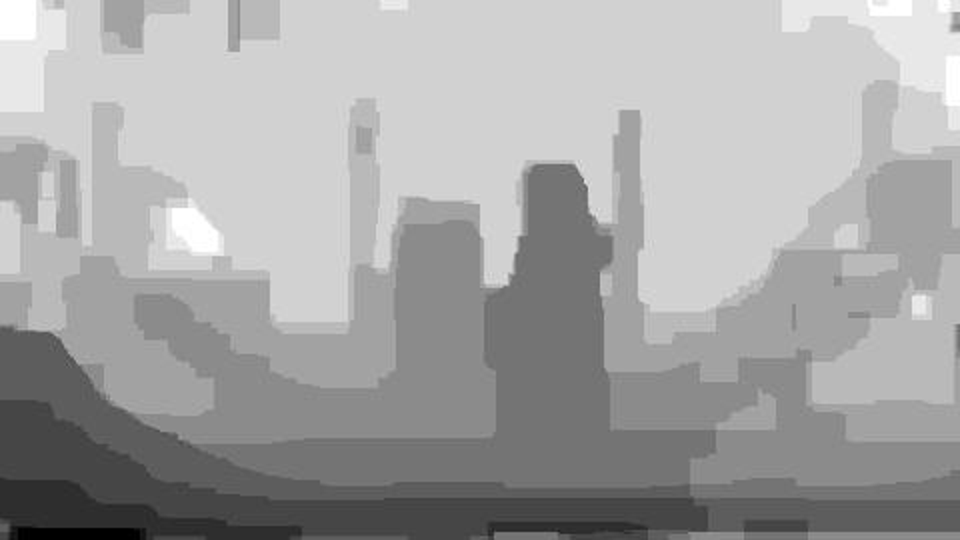}
        \includegraphics[width=1\linewidth]{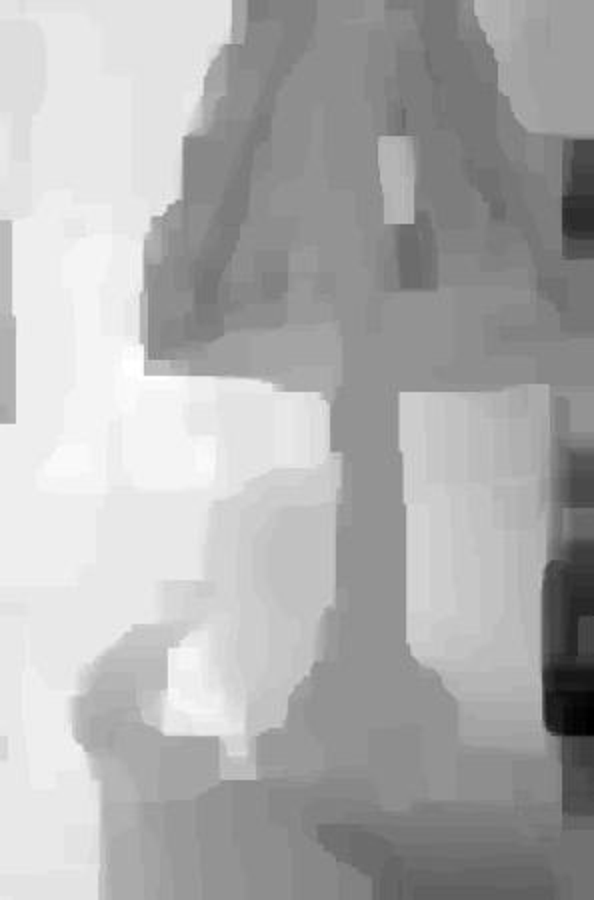}
        \includegraphics[width=1\linewidth]{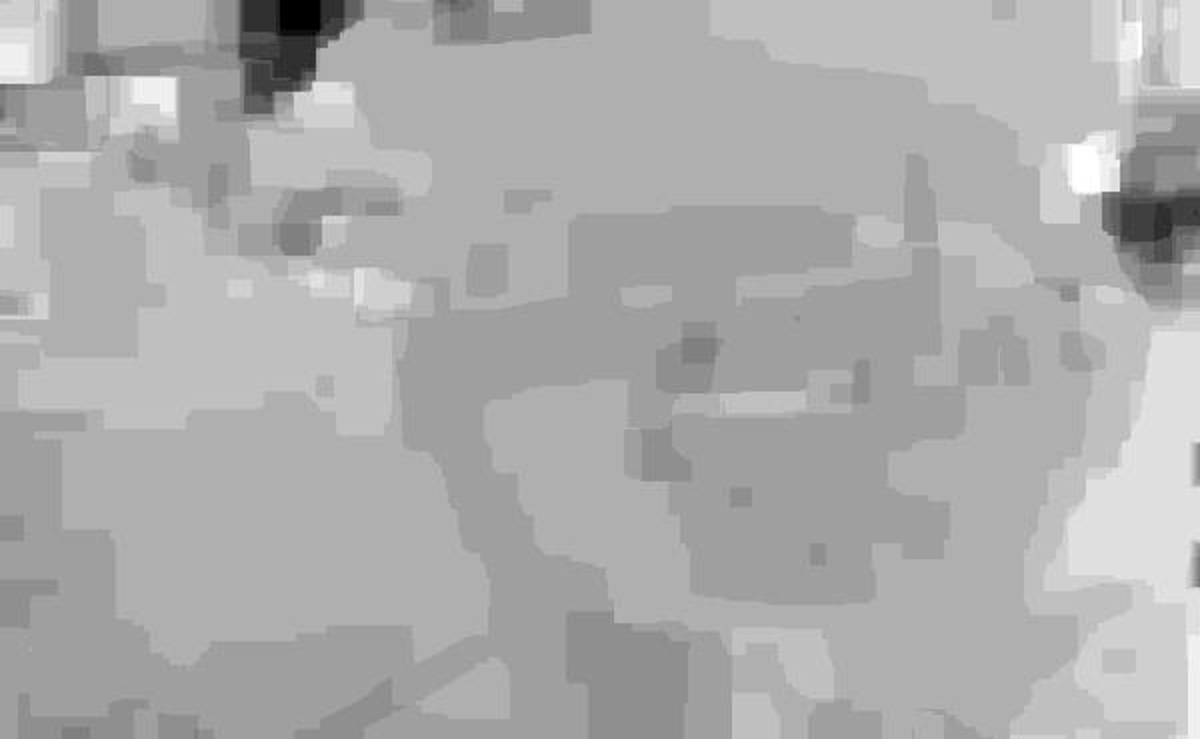}
        Depth
    \end{minipage}
    \begin{minipage}[t]{0.1\textwidth}
    \vspace{0pt}
        \centering
        \includegraphics[width=1\linewidth]{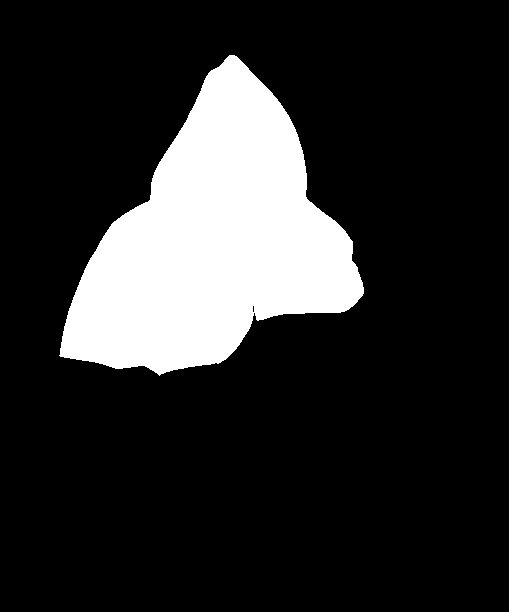}
        \includegraphics[width=1\linewidth]{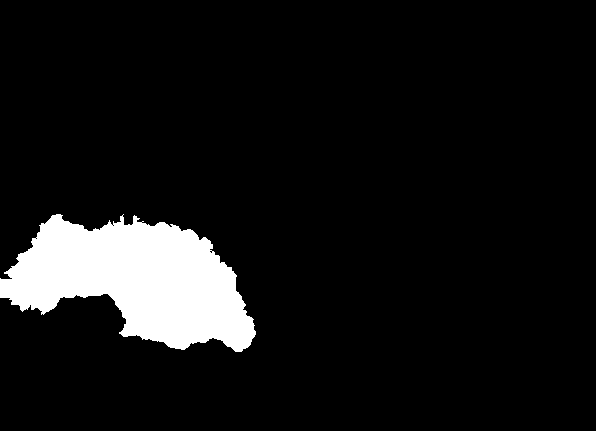}
        \includegraphics[width=1\linewidth]{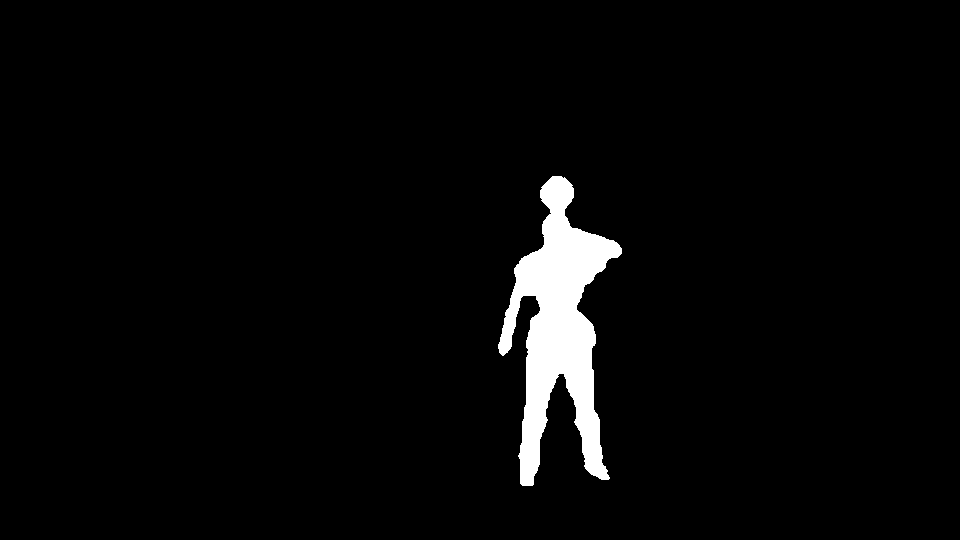}
        \includegraphics[width=1\linewidth]{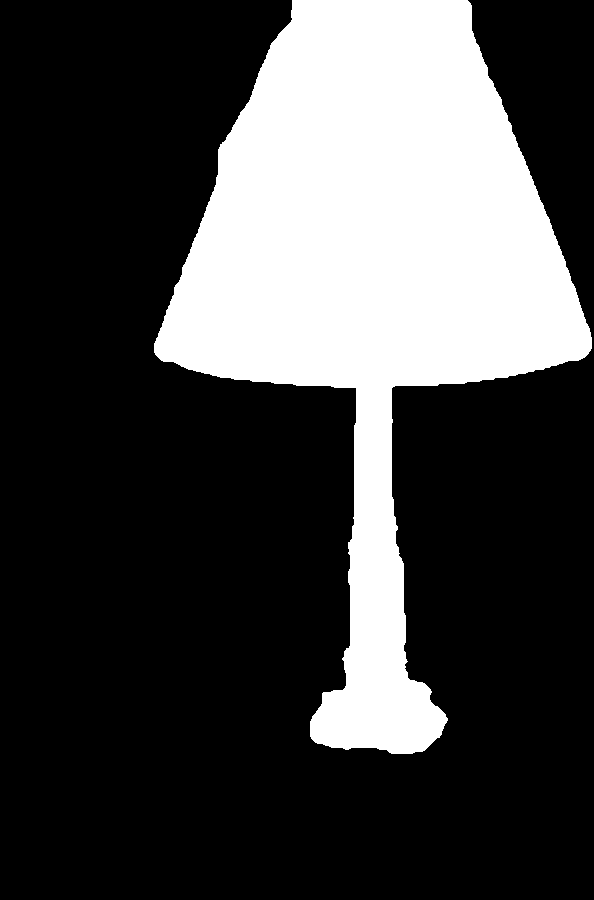}
        \includegraphics[width=1\linewidth]{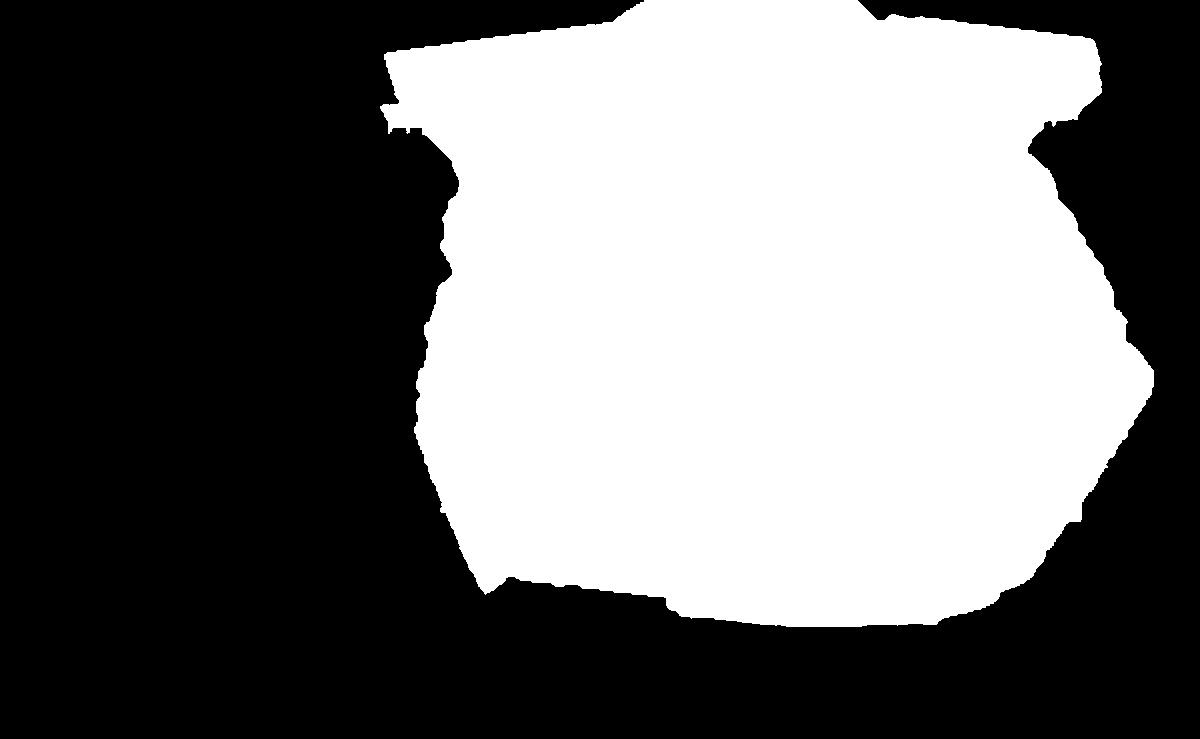}
        \textbf{GT}
    \end{minipage}
    \begin{minipage}[t]{0.1\textwidth}
    \vspace{0pt}
        \centering
        \includegraphics[width=1\linewidth]{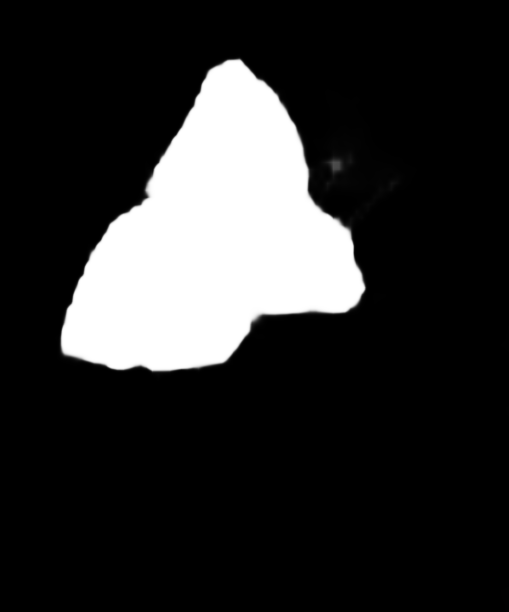}
        \includegraphics[width=1\linewidth]{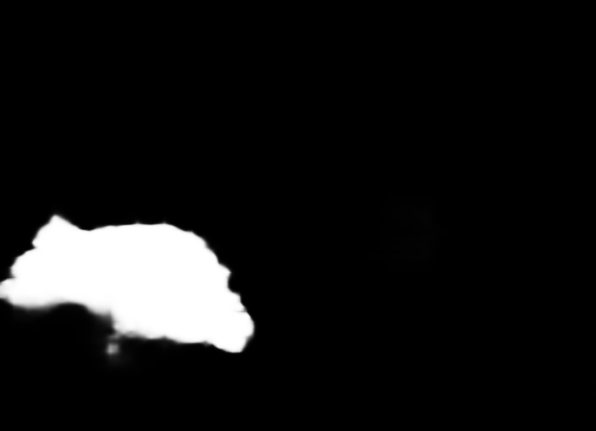}
        \includegraphics[width=1\linewidth]{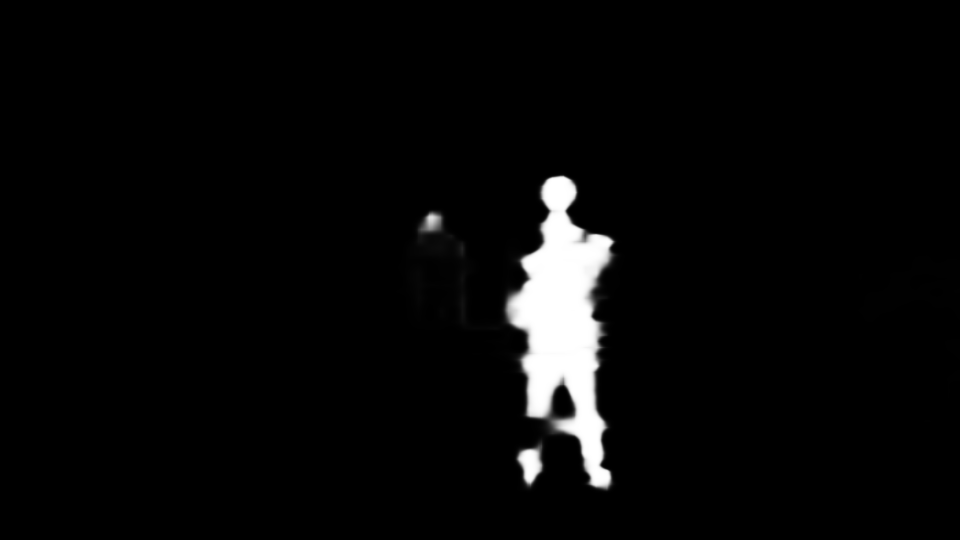}
        \includegraphics[width=1\linewidth]{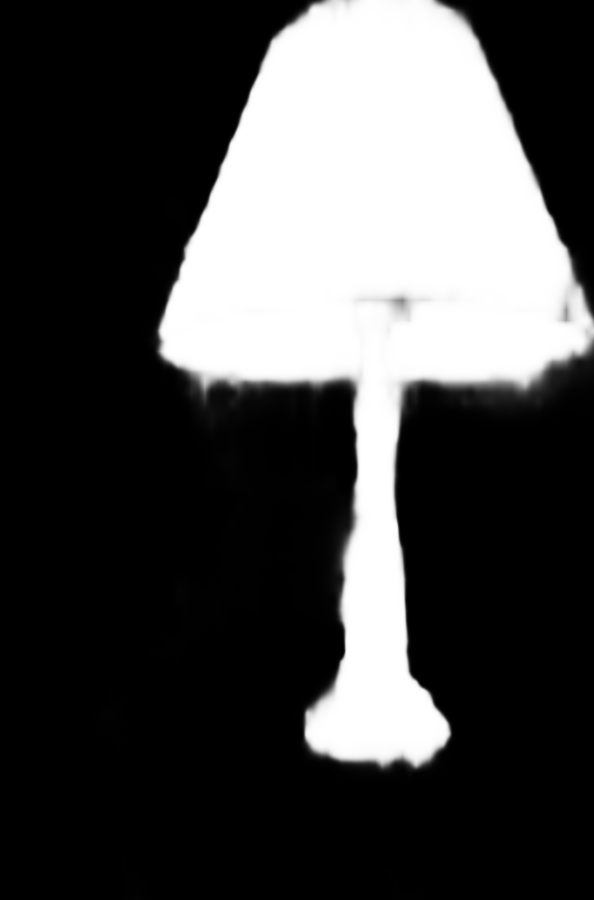}
        \includegraphics[width=1\linewidth]{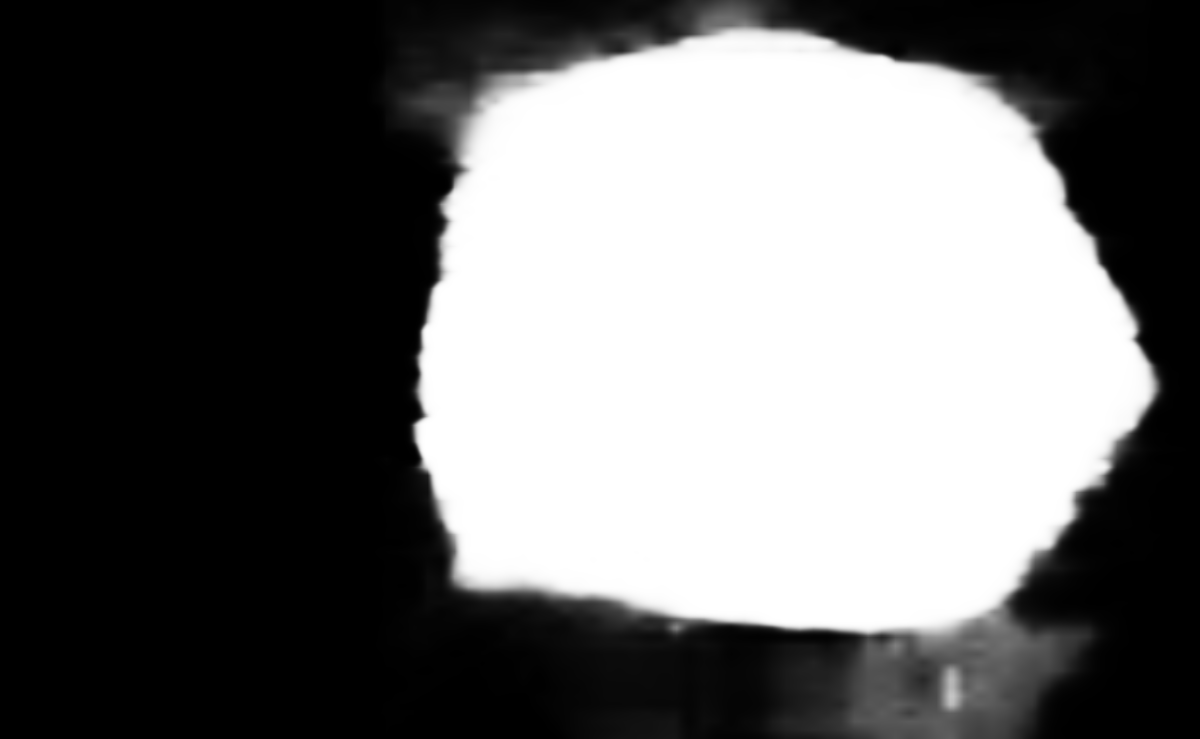}
        \textbf{Ours}
    \end{minipage}
    \begin{minipage}[t]{0.1\textwidth}
    \vspace{0pt}
        \centering
        \includegraphics[width=1\linewidth]{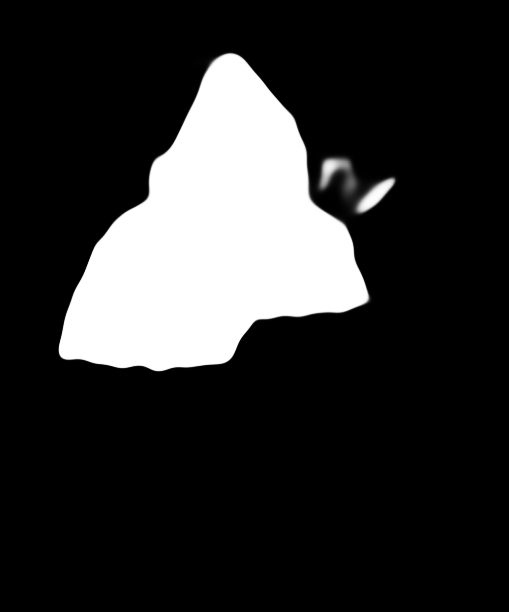}
        \includegraphics[width=1\linewidth]{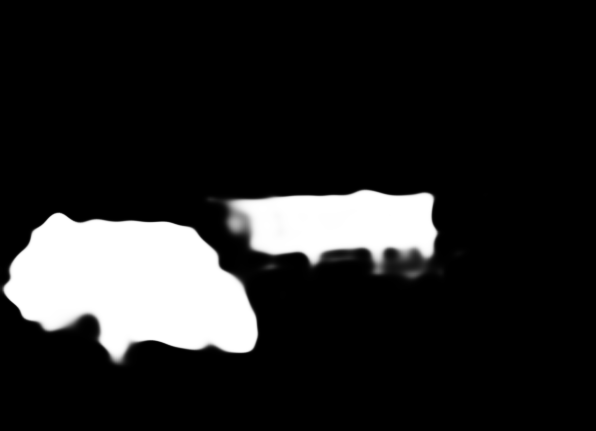}
        \includegraphics[width=1\linewidth]{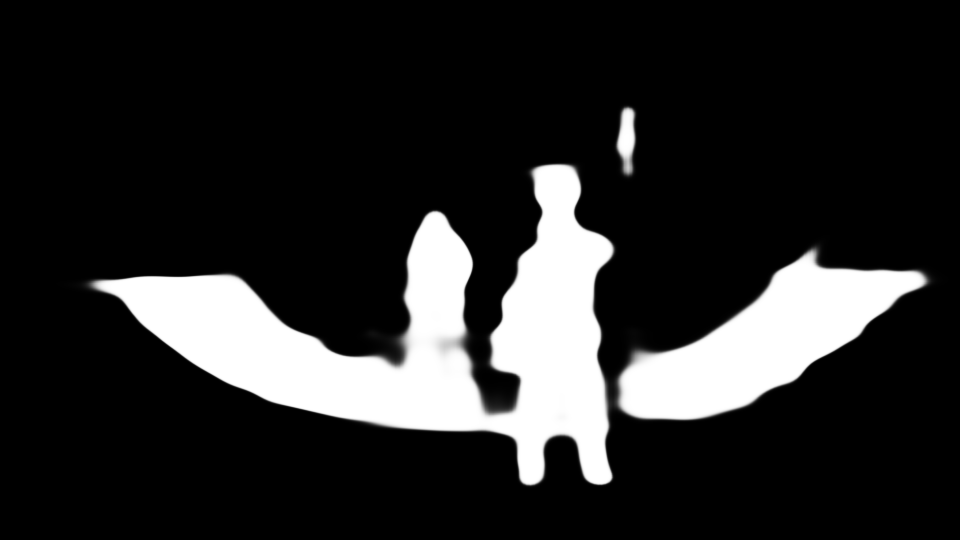}
        \includegraphics[width=1\linewidth]{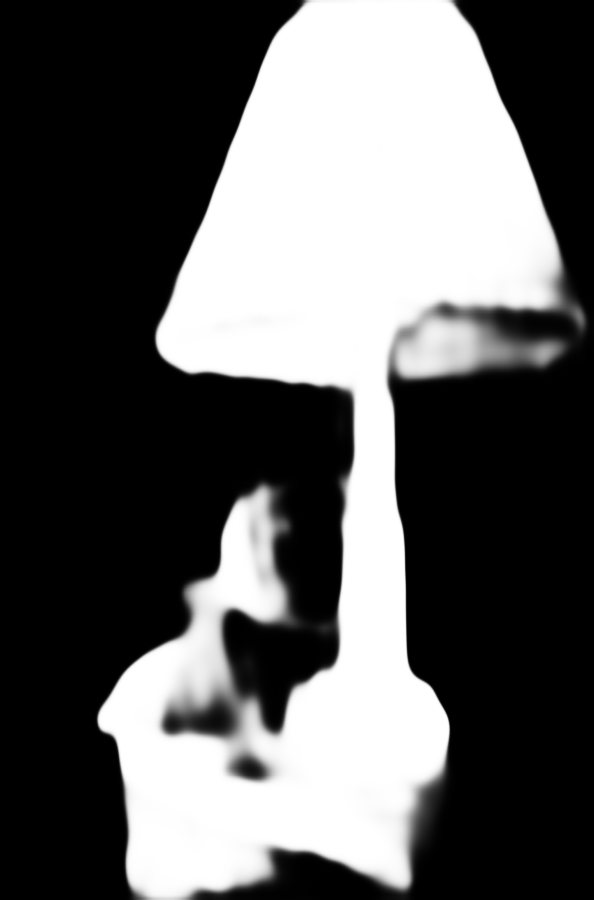}
        \includegraphics[width=1\linewidth]{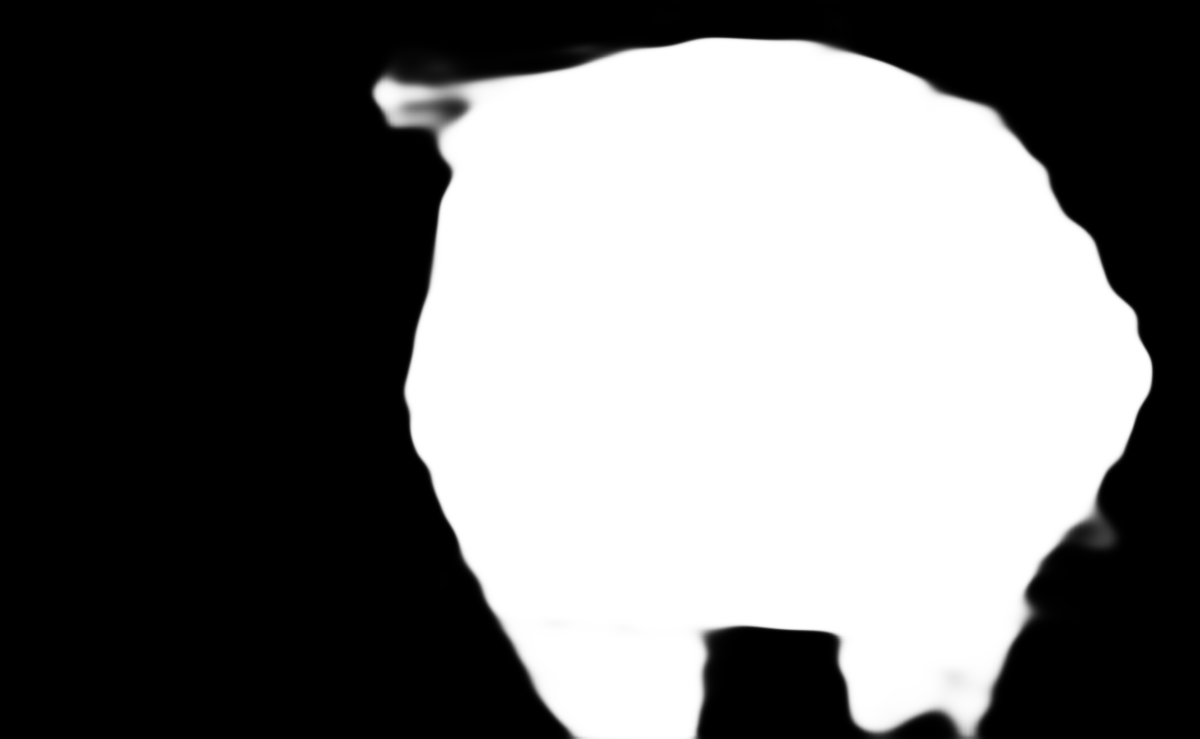}
        TriT\cite{tritransnet}
    \end{minipage}
    \begin{minipage}[t]{0.1\textwidth}
    \vspace{0pt}
        \centering
        \includegraphics[width=1\linewidth]{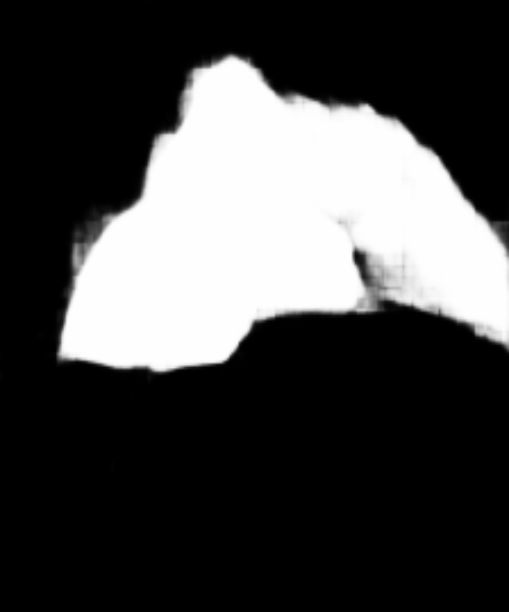}
        \includegraphics[width=1\linewidth]{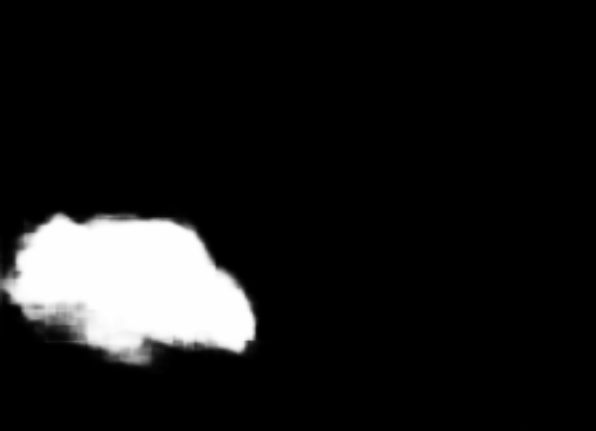}
        \includegraphics[width=1\linewidth]{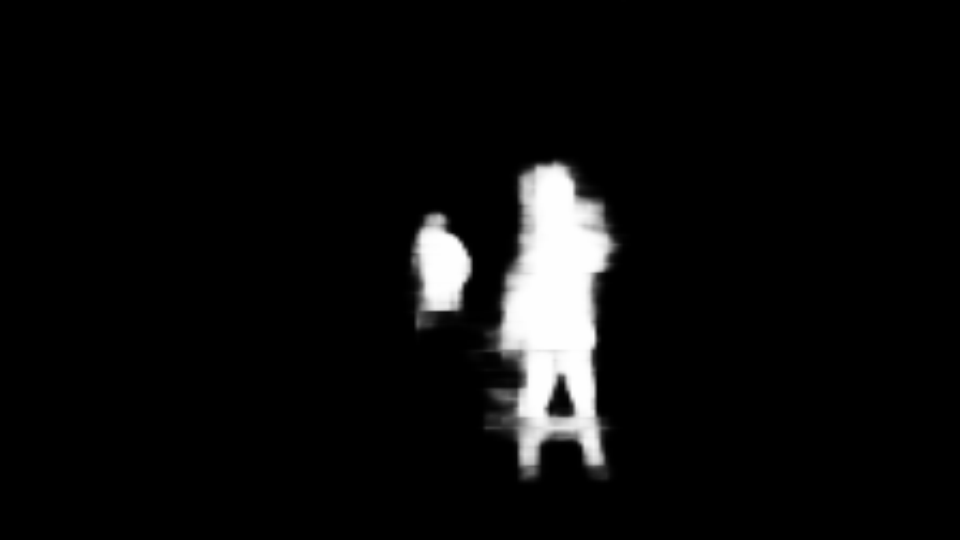}
        \includegraphics[width=1\linewidth]{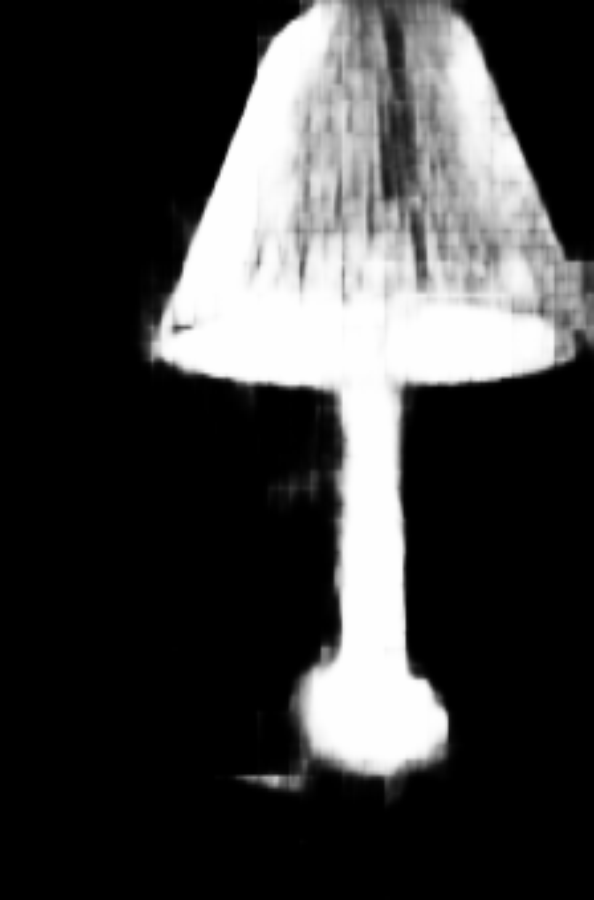}
        \includegraphics[width=1\linewidth]{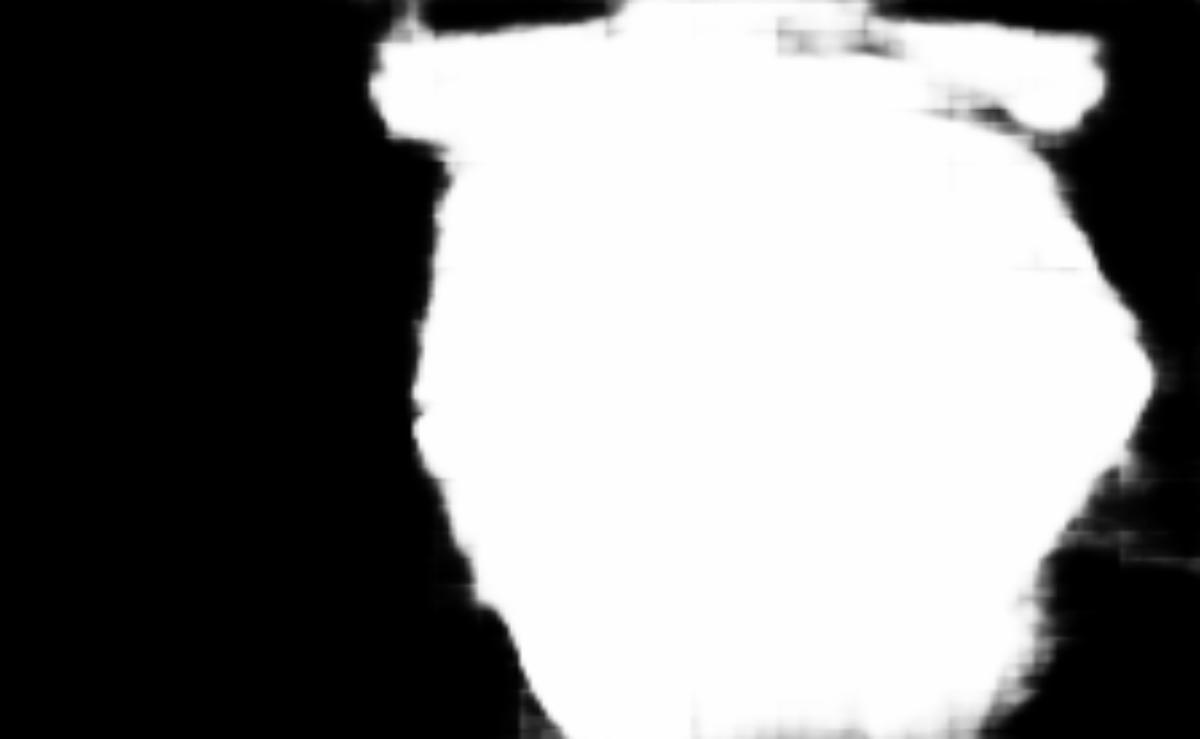}
        VST\cite{vst}
    \end{minipage}
    \begin{minipage}[t]{0.1\textwidth}
    \vspace{0pt}
        \centering
        \includegraphics[width=1\linewidth]{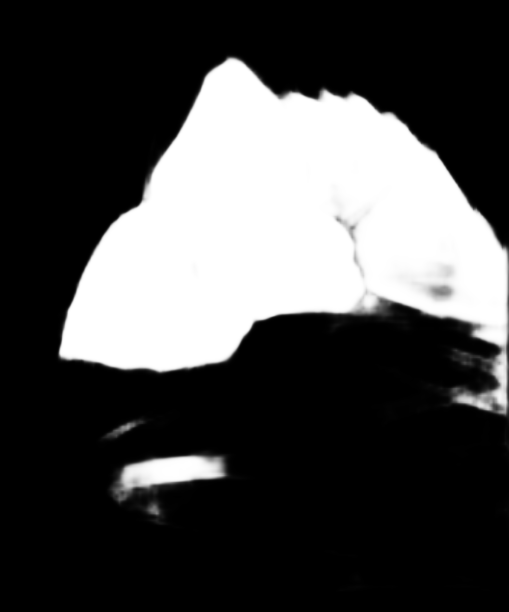}
        \includegraphics[width=1\linewidth]{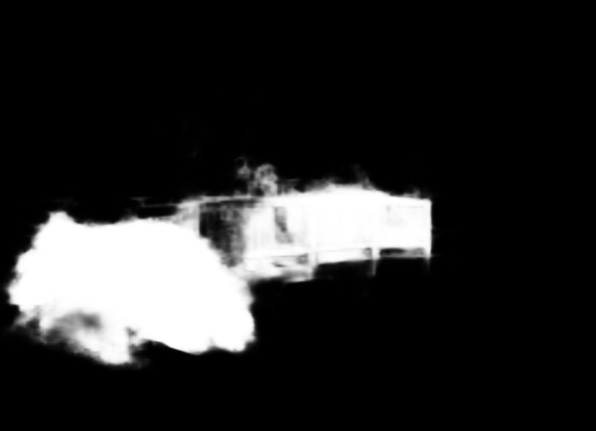}
        \includegraphics[width=1\linewidth]{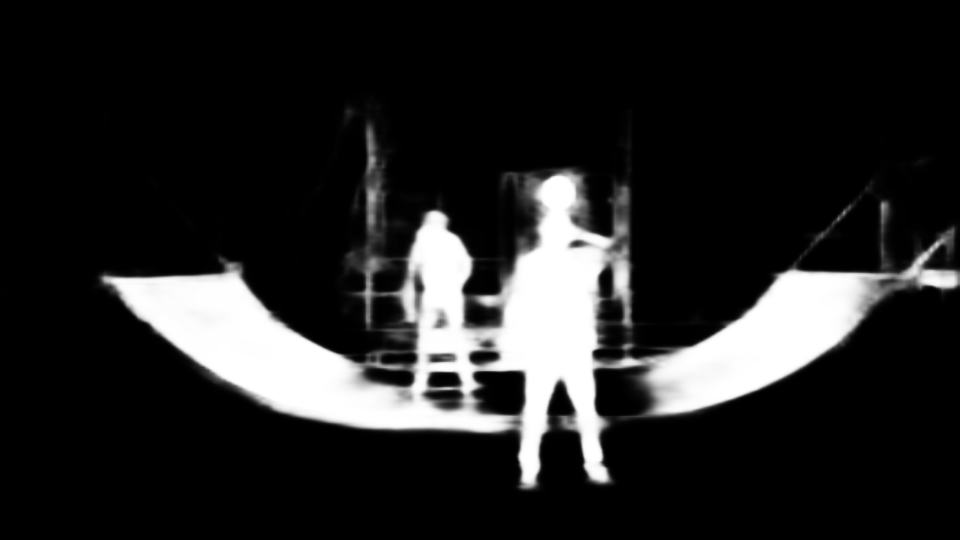}
        \includegraphics[width=1\linewidth]{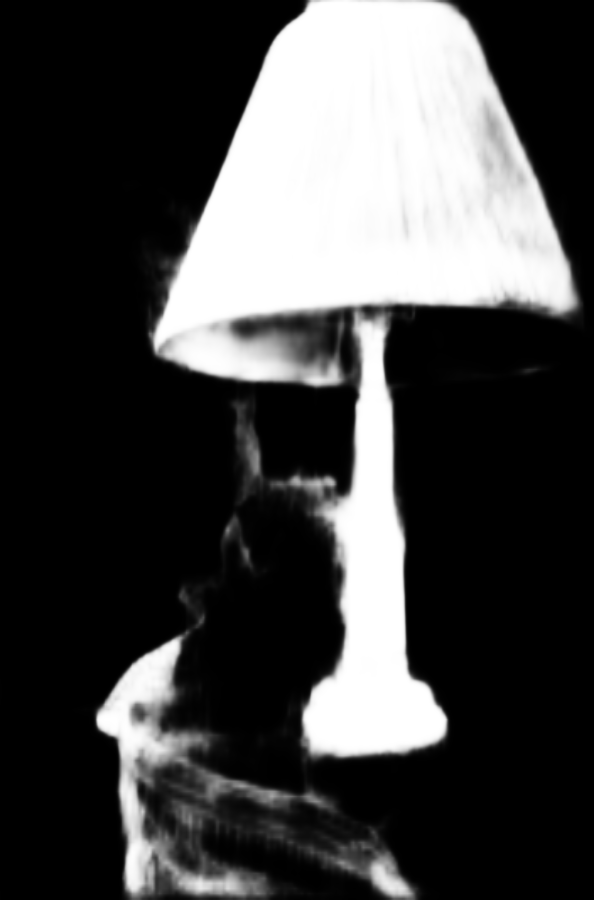}
        \includegraphics[width=1\linewidth]{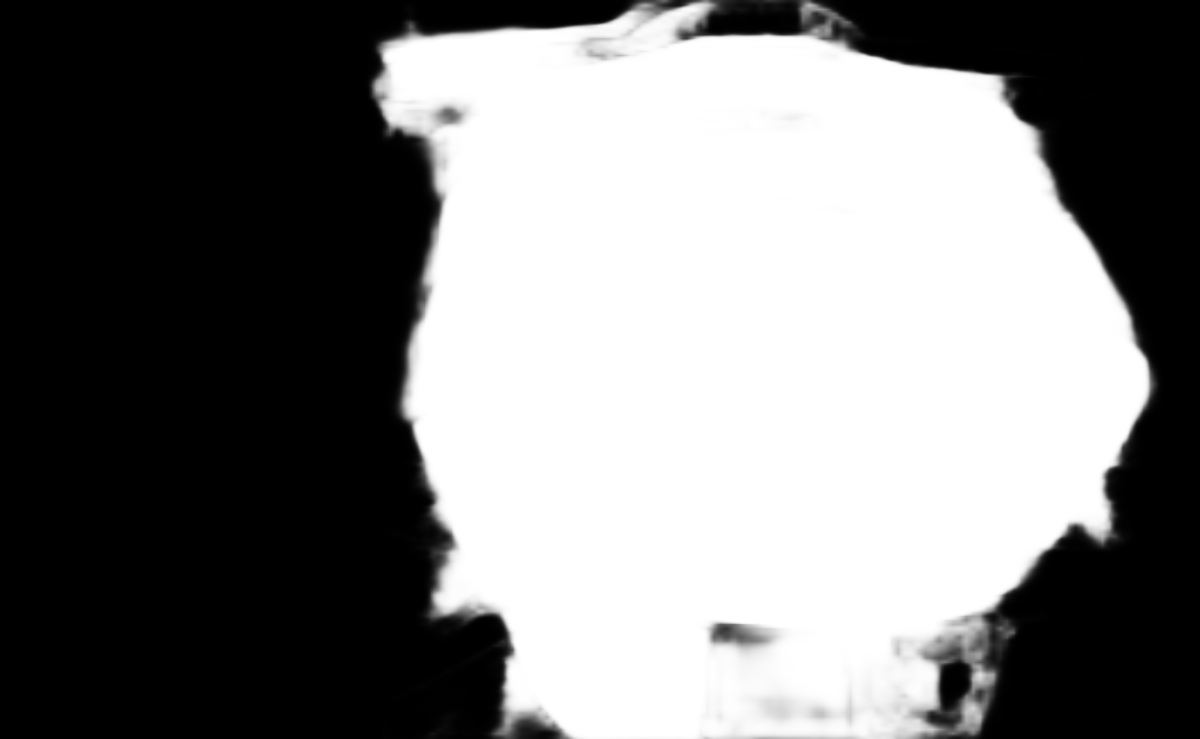}
        RD3D\cite{rd3d}
    \end{minipage}
    \begin{minipage}[t]{0.1\textwidth}
    \vspace{0pt}
        \centering
        \includegraphics[width=1\linewidth]{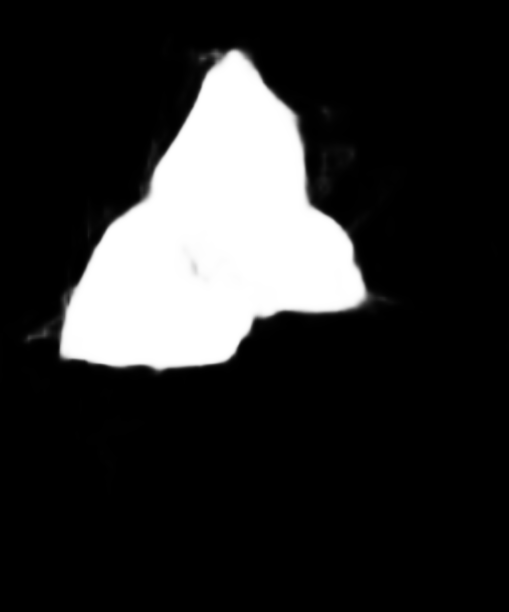}
        \includegraphics[width=1\linewidth]{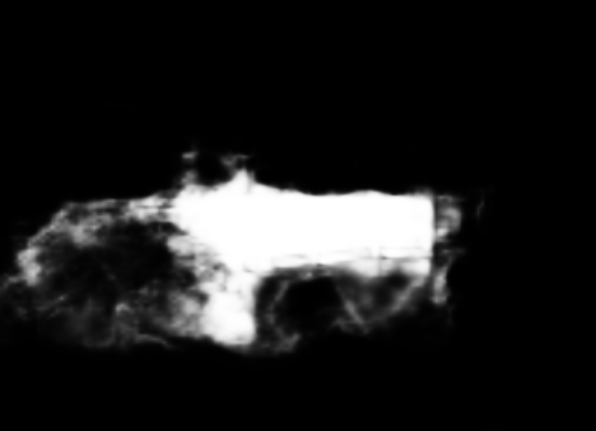}
        \includegraphics[width=1\linewidth]{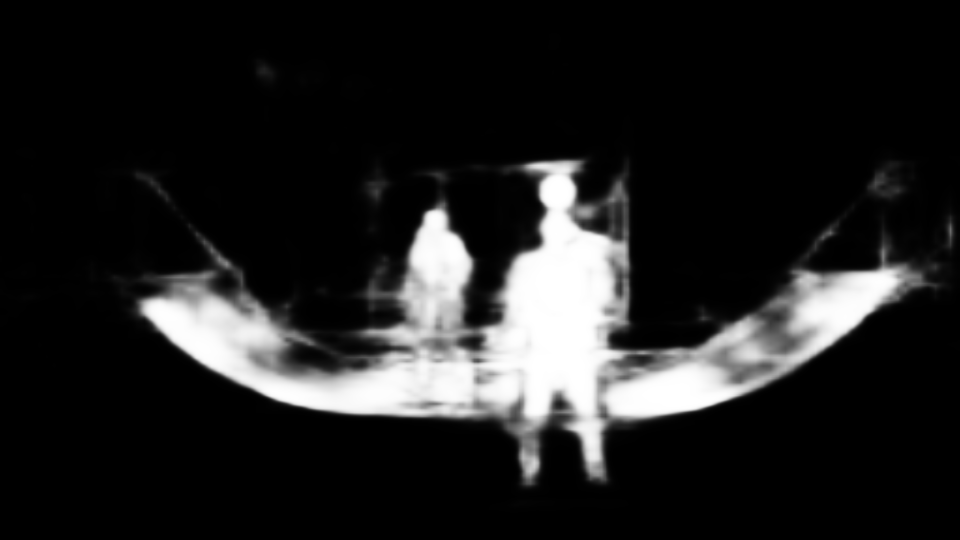}
        \includegraphics[width=1\linewidth]{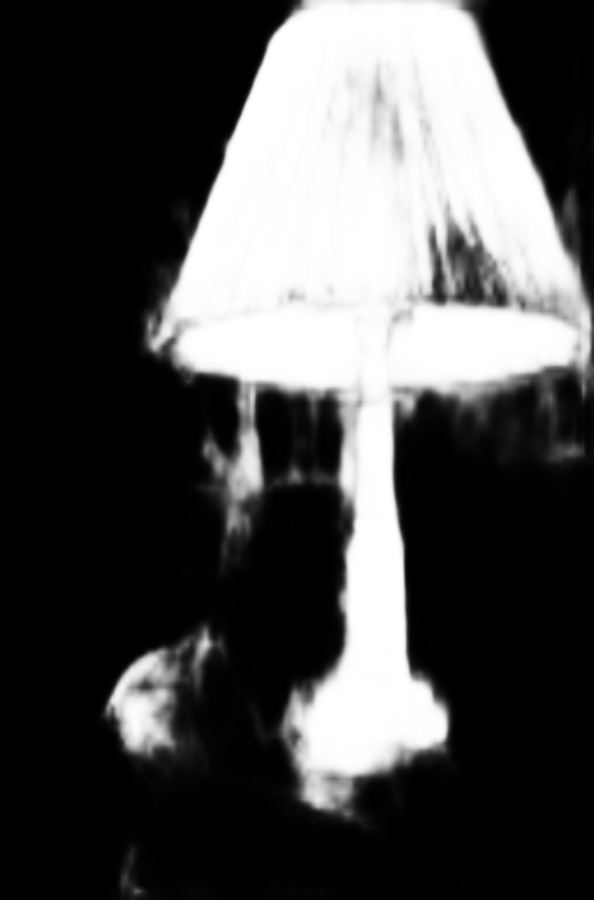}
        \includegraphics[width=1\linewidth]{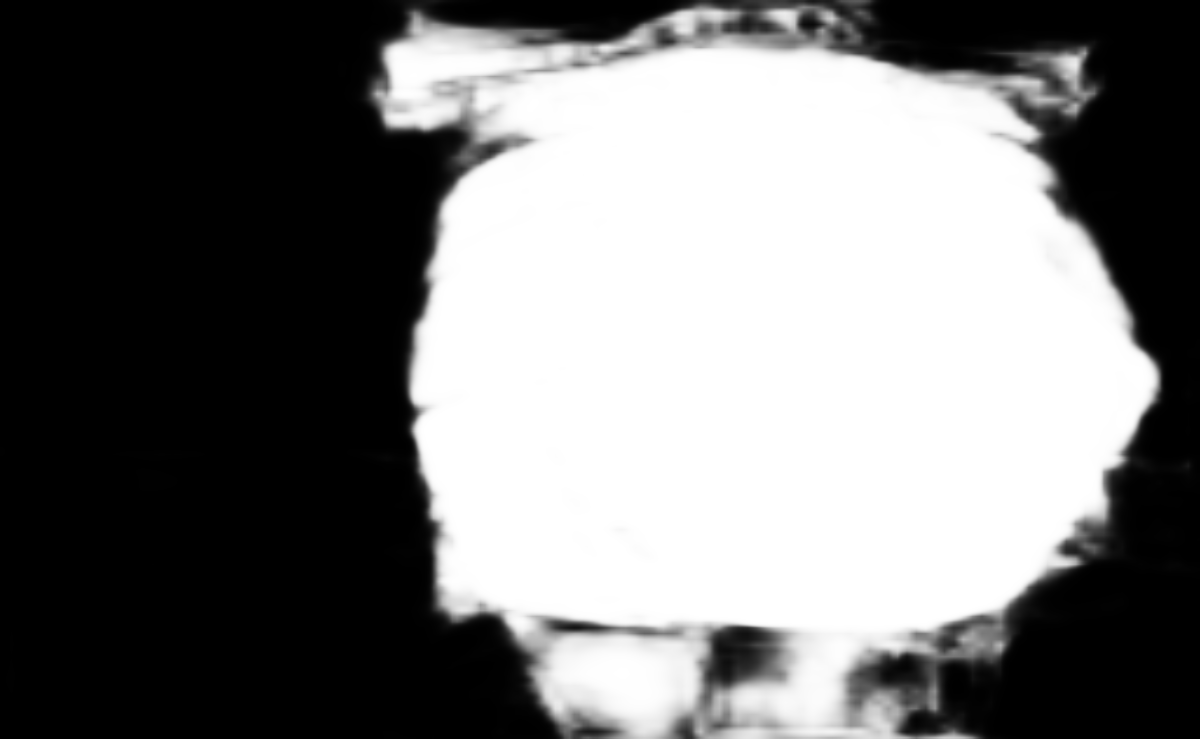}
        DFM\cite{dfmnet}
    \end{minipage}
    \begin{minipage}[t]{0.1\textwidth}
    \vspace{0pt}
        \centering
        \includegraphics[width=1\linewidth]{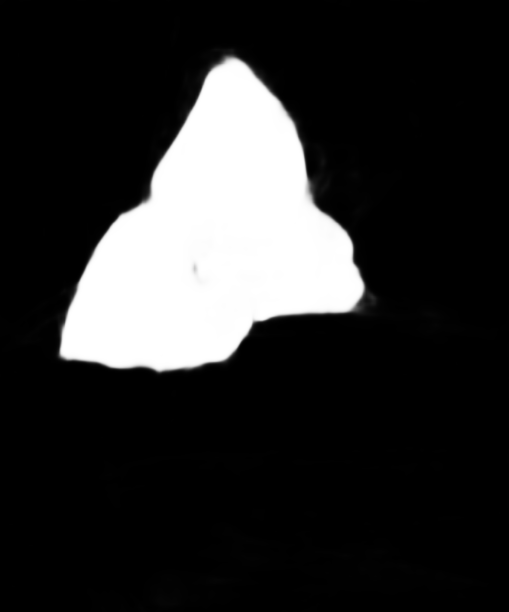}
        \includegraphics[width=1\linewidth]{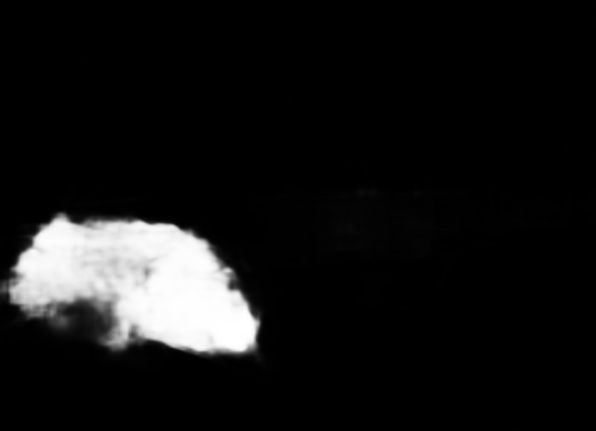}
        \includegraphics[width=1\linewidth]{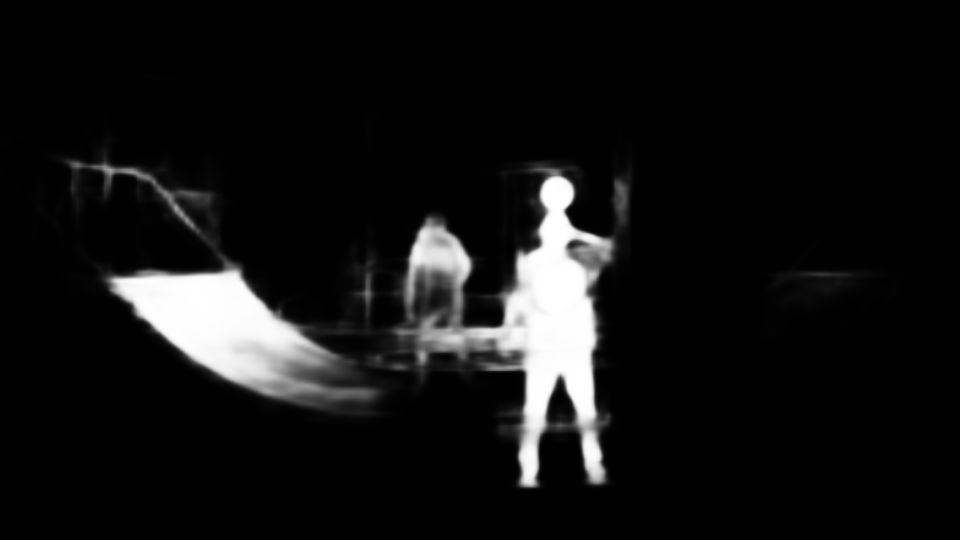}
        \includegraphics[width=1\linewidth]{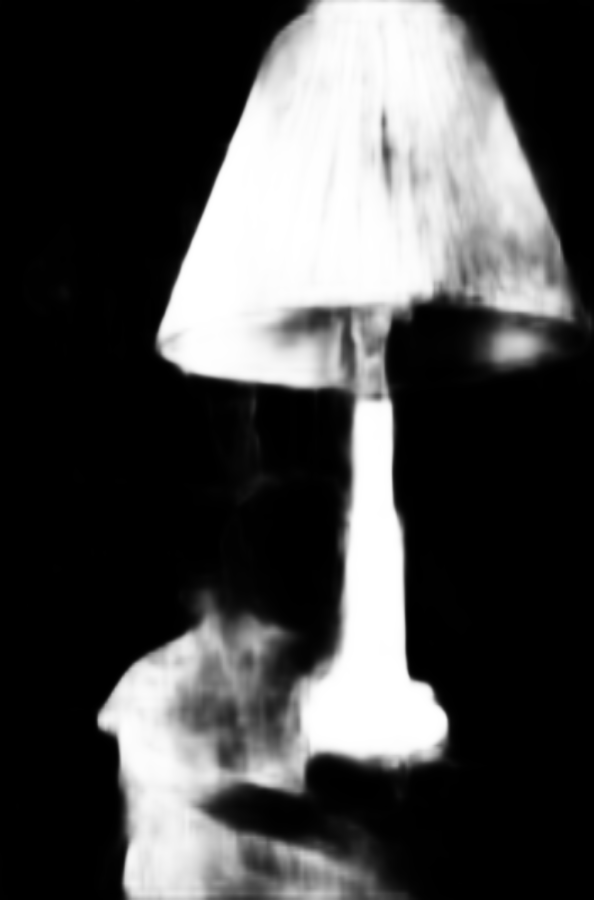}
        \includegraphics[width=1\linewidth]{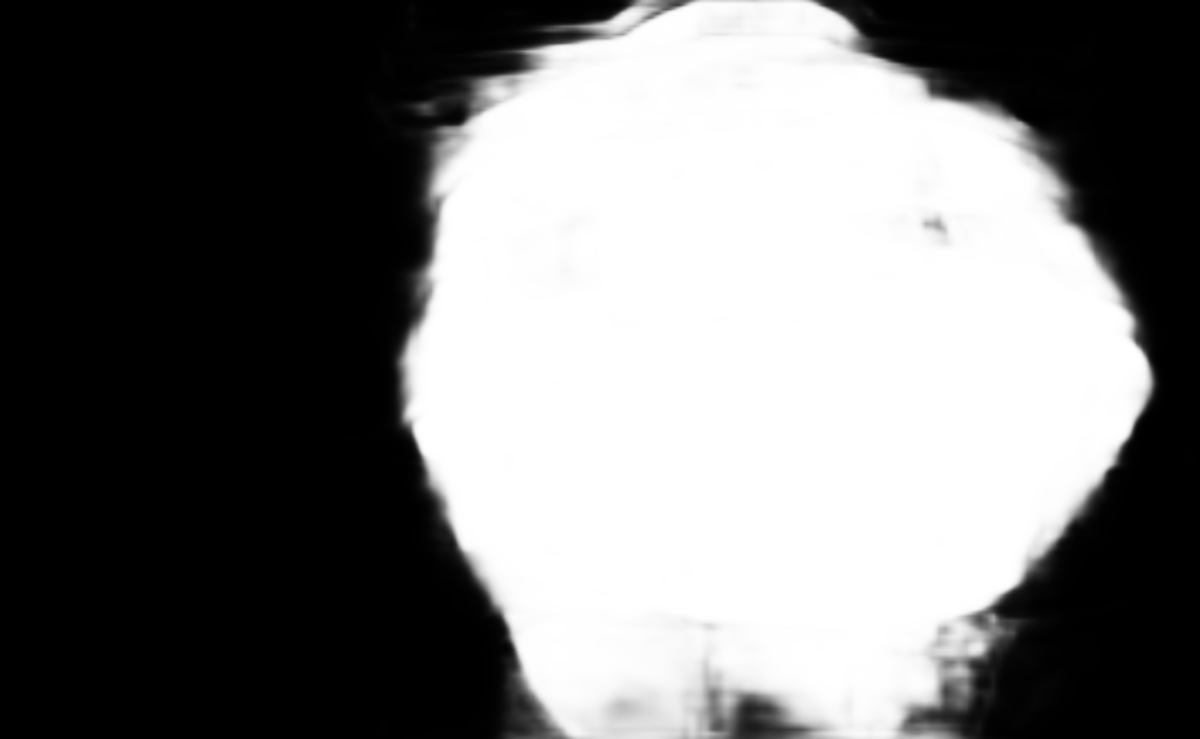}
        BTS\cite{btsnet}
    \end{minipage}
    \caption{Qualitative comparison of our proposed method against the state-of-the-art \textbf{RGB-D} SOD methods.}
    \label{fig:visual_rgbd}
\end{figure*}

We involve 6 traditional methods, 16 CNN-based methods and 2 newly developed Transformer-based methods (TriTransNet~\cite{tritransnet} and VST~\cite{vst}) for performance comparison. The SOD performances of different SOTA approaches and our DFTR are presented in Table~\ref{tab:rgbd}. It can be observed that our DFTR achieves the best SOD performances under most metrics on different datasets. Especially, on NLPR and SSD datasets, our DFTR consistently outperforms all the listed benchmarking frameworks, which validates its superiority. Our DFTR outperforms Transformer-based methods (\emph{i.e.}, VST and TriTransNet) on 16 of 20 metrics, while slightly worse than VST and TriTransNet on the rest 4 metrics. The underlying reason is that those approaches adopt depth maps as the auxiliary input to assist salient object detection. In contrast, the proposed DFTR relaxes the requirement of depth maps for model inference, and only takes the RGB images as input.
We further visualize the saliency maps predicted by our model and other methods for qualitative comparison. As illustrated in Fig.~\ref{fig:visual_rgbd}, our DFTR generates more accurate and clearer saliency maps compared to other methods.
\begin{table*}[t]
  \centering
  \renewcommand{\arraystretch}{1.05} 
  \renewcommand{\tabcolsep}{1mm} 
  \caption{Quantitative comparison of the proposed DFTR method with 24 \textbf{RGB-D SOD} models (including 6 traditional models, 16 CNN-based  models and 2 Transformer-based models) on 5 public RGB-D saliency benchmark datasets in terms of 4 widely used evaluation metrics (\emph{i.e.}, structure measure $S_{\alpha}$ \cite{fan2017structure}, maximum F-measure $F_{\beta}$ \cite{achanta2009frequency}, maximum enhanced alignment measure $E_{\xi}$\cite{Fan2018Enhanced}, and mean absolute error $M$ \cite{perazzi2012saliency}). 
  ``$\uparrow$'' (or ``$\downarrow$'') indicates that larger (or smaller) is better. The subscript of each model denotes the publication year. 
  In each column, the best result is marked with \red{red} and the second best with \blue{blue}. 
  }
  \label{tab:rgbd}
  
  \vspace{0.15mm}
  \scriptsize
  \resizebox{\textwidth}{!}{\begin{tabular}{l|cccc|cccc|cccc|cccc|cccc}
  \hline\toprule
    \multicolumn{1}{c|}{\multirow{2}{*}{\textbf{Model}}} 
     &\multicolumn{4}{c|}{NJU2K~\cite{ju2014depth}}
    &\multicolumn{4}{c|}{STERE~\cite{niu2012leveraging}}
    &\multicolumn{4}{c|}{NLPR~\cite{peng2014rgbd}}
    &\multicolumn{4}{c|}{SSD~\cite{zhu2017three}}
    &\multicolumn{4}{c}{SIP~\cite{fan2019rethinking}}\\

     &$S_{\alpha}\uparrow$   &$F_{\beta}\uparrow$    &$E_{\xi}\uparrow$  &$M\downarrow$  
    &$S_{\alpha}\uparrow$   &$F_{\beta}\uparrow$    &$E_{\xi}\uparrow$  &$M\downarrow$  
    &$S_{\alpha}\uparrow$   &$F_{\beta}\uparrow$    &$E_{\xi}\uparrow$  &$M\downarrow$  
    &$S_{\alpha}\uparrow$   &$F_{\beta}\uparrow$    &$E_{\xi}\uparrow$  &$M\downarrow$   
    &$S_{\alpha}\uparrow$   &$F_{\beta}\uparrow$    &$E_{\xi}\uparrow$  &$M\downarrow$ \\
   
  \midrule
     LBE$_{16}$~\cite{feng2016local}
    & .695   & .748   & .803   & .153
    & .660   & .633   & .787   & .250
    & .762   & .745   & .855   & .081
    & .621   & .619   & .736   & .278
    & .727   & .751   & .853   & .200	\\    

     DCMC$_{16}$~\cite{cong2016saliency}
     & .686   & .715   & .799   & .172
    & .731   & .740   & .819   & .148
    & .707   & .666   & .773   & .111
    & .704   & .711   & .786   & .169
    & .683   & .618   & .743   & .186	\\       

     SE$_{16}$~\cite{guo2016salient}
    & .664   & .748   & .813   & .169
    & .708   & .755   & .846   & .143
    & .756   & .713   & .847   & .091
    & .675   & .710   & .800   & .165
    & .628   & .661   & .771   & .164	\\

     MDSF$_{17}$~\cite{song2017depth}
    & .748   & .775   & .838   & .157
    & .728   & .719   & .809   & .176
    & .805   & .793   & .885   & .095
    & .673   & .703   & .779   & .192
    & .717   & .698   & .798   & .167	\\

     CDCP$_{17}$~\cite{zhu2017innovative}
    & .669 & .621 & .741 & .180
    & .713 & .664 & .786 & .149
    & .669 & .621 & .741 & .180
    & .603 & .535 & .700 & .214
    & .595 & .505 & .721 & .224 \\
    
     DTM$_{20}$~\cite{cong2019going}
    & .706 & .716 & .799 & .190
    & .747 & .743 & .837 & .168
    & .733 & .677 & .833 & .145
    & .677 & .651 & .773 & .199
    & .690 & .659 & .778 & .203
	\\   
    \midrule
      
     ICNet$_{20}$~\cite{li2020icnet}
    & .894   & .891   & .926   & .052
    & .903   & .898   & .942   & .045
    & .923   & .908   & .952   & .028
    & .848   & .841   & .902   & .064
    & .854   & .857   & .903   & .069	\\
     S${^2}$MA$_{20}$~\cite{liu2020}
    & .894   & .889   & .930   & .053
    & .890   & .882   & .932   & .051
    & .915   & .902   & .953   & .030
    & .868   & .848   & .909   & .052
    & .872   & .877   & .919   & .057	\\    
     A2dele$_{20}$~\cite{piao2020}
    & .871   & .874   & .916   & .051
    & .878   & .879   & .928   & .044
    & .898   & .882   & .944   & .029
    & .802   & .776   & .861   & .070
    & .828   & .833   & .889   & .070	\\

     SSF$_{20}$~\cite{zhang2020}
    & .899   & .896   & .935   & .043
    & .893   & .890   & .936   & .044
    & .914   & .896   & .953   & .026
    & .845   & .824   & .897   & .058
    & .876   & .882   & .922   & .052	\\
    
     UCNet$_{20}$~\cite{zhang2020uc}
    & .897   & .895   & .936   & .043
    & .903   & .899   & .944   & .039
    & .920   & .903   & .956   & .025
    & .865   & .854   & .907   & .049
    & .875   & .879   & .919   & .051	\\

     Cas-GNN$_{20}$~\cite{luoECCV2020}
    & .911   & .903   & .933   & .035
    & .899   & .901   & .930   & .039
    & .919   & .904   & .947   & .028
    & .872   & .862   & .915   & .047
    & .875   & .879   & .919   & .051	\\
    
     CMMS$_{20}$~\cite{li2020}
    & .900   & .897   & .936   & .044
    & .895   & .893   & .939   & .043
    & .915   & .896   & .949   & .027
    & .874   & .864   & .922   & .046
    & .872   & .877   & .911   & .058	\\

     CoNet$_{20}$~\cite{Wei_2020_ECCV}
    & .895   & .893   & .937   & .046
    & .908   & .905   & .949   & .040
    & .908   & .887   & .945   & .031
    & .853   & .840   & .915   & .059
    & .858   & .867   & .913   & .063	\\
    
    DANet$_{20}$~\cite{zhaoeccv20}
    & .899   & .910   & .935   & .045
    & .901   & .892   & .937   & .043
    & .915   & .916   & .953   & .028
    & .864   & .866   & .914   & .050
    & .875   & .892   & .918   & .054	\\
    
    DASNet$_{20}$~\cite{dasnet}
    & .902   & .902   & .939   & .042
    & .910   & .904   & .944   & .037
    & .929   & .922   & .964   & .021
    & .885   & .872   & .930   & .042
    & .877   & .886   & .925   & .051	\\

     BBS-Net$_{20}$~\cite{bbsnet}
    & .921   & .920   & .949   & .035
    & .908   & .903   & .942   & .041
    & .930   & .918   & .961   & .023
    & .882   & .859   & .919   & .044
    & .879   & .883   & .922   & .055	\\

    D$^{3}$Net$_{21}$~\cite{fan2019rethinking}
    & .900   & .900   & .950   & .041
    & .899   & .891   & .938   & .046
    & .912   & .897   & .953   & .030
    & .857   & .834   & .910   & .058
    & .860   & .861   & .909   & .063	\\

    JLDCF$_{21}$~\cite{jldcf}
    & .911   & .913   & .948   & .040
    & .911   & .907   & .949   & .039
    & .926   & .917   & .964   & .023
    & - &-&-&-
    & .892   & .900   & \red{.949}   & .046	\\
        
    DFMNet$_{21}$~\cite{dfmnet}
    & .906   & .910   & .947   & .042
    & .898   & .893   & .941   & .045
    & .923   & .908   & .957   & .026
    & - &-&-&-
    & .883   & .887   & .926   & .051	\\
    
    RD3D$_{21}$~\cite{rd3d}
    & .916   & .914   & .947   & .036
    & .911   & .906   & .947   & .037
    & .930   & .919   &\blue{ .965}   & .022
    & - &-&-&-
    & .885   & .889   & .924   & .048	\\
    
    BTSNet$_{21}$~\cite{btsnet}
    & .921   & \red{.924}   & \blue{.954}   & .036
    & \blue{.915}   & \blue{.911}   & .949   & .038
    & \blue{.934}   & \blue{.923}   &\blue{.965}   & .023
    & - &-&-&-
    & .896   & .901   & .933   & .044	\\ 
    
    \midrule
    TriTransNet$_{21}$~\cite{tritransnet}
    & .920   & .919   & \red{.960}   & \red{.020}
    & .908   & .893   & .927   & \red{.033}
    & .928   & .909   & .960   & \blue{.020}
    & -   & -   & -   & -
    & .886   & .892   & .924   & .043	\\
    VST$_{21}$~\cite{vst}
    & \red{.922}   & .920   & .951   & .035
    & .913   & .907   & \red{.951}   & .038
    & .932   & .920   & .962   & .024
    & \blue{.889}   & \blue{.876}   & \blue{.935}   & \blue{.045}
    & \red{.904}   & \red{.915}   & .944  & \red{.040}	\\
    
    \midrule

    \textbf{DFTR~(Ours)}
    & \red{.922} & \blue{.923} & \blue{.954} & \blue{.034}
    & \red{.918} & \red{.914} & \red{.951} & \blue{.034}
    & \red{.941} & \red{.934} & \red{.972} & \red{.018}
    &\red{.890} & \red{.882} & \red{.937} & \red{.036}
    & \red{.904} & \blue{.913} & \blue{.946} & \red{.040} \\
    \bottomrule
    \hline
        
  \end{tabular}
  }
\end{table*}

\subsubsection{RGB SOD comparison.} 
\begin{figure*}[htbp]
    \scriptsize
    \centering
    \begin{minipage}[t]{0.1\textwidth}
    \vspace{0pt}
        \centering
        \includegraphics[width=1\linewidth]{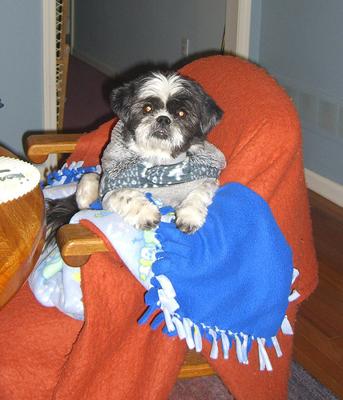}
        \includegraphics[width=1\linewidth]{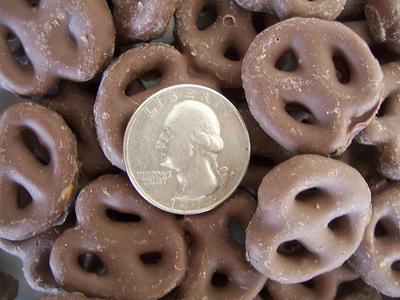}
        \includegraphics[width=1\linewidth]{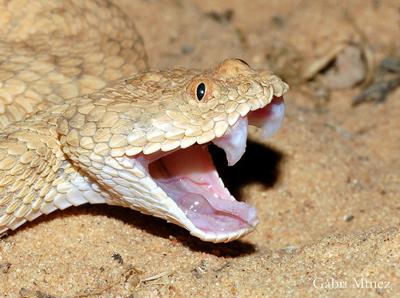}
        \includegraphics[width=1\linewidth]{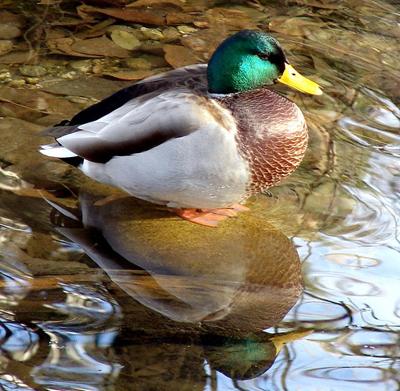}
        \includegraphics[width=1\linewidth]{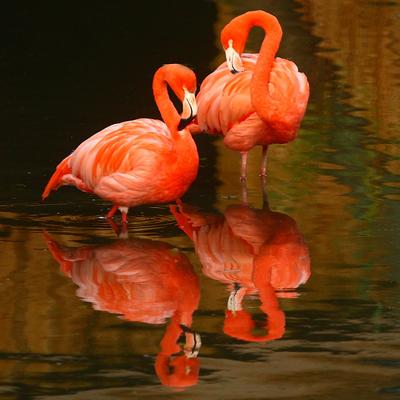}
        RGB
    \end{minipage}
    \begin{minipage}[t]{0.1\textwidth}
    \vspace{0pt}
        \centering
        \includegraphics[width=1\linewidth]{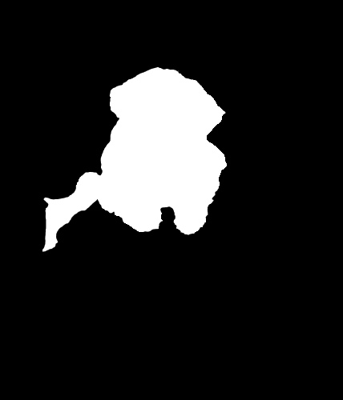}
        \includegraphics[width=1\linewidth]{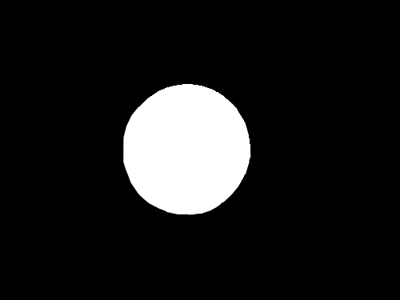}
        \includegraphics[width=1\linewidth]{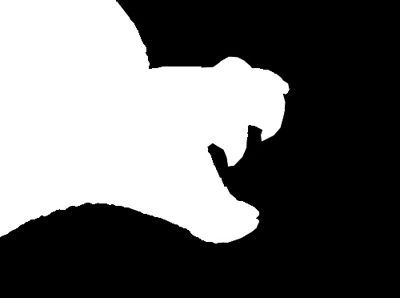}
        \includegraphics[width=1\linewidth]{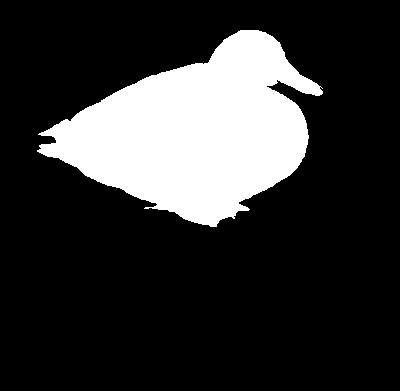}
        \includegraphics[width=1\linewidth]{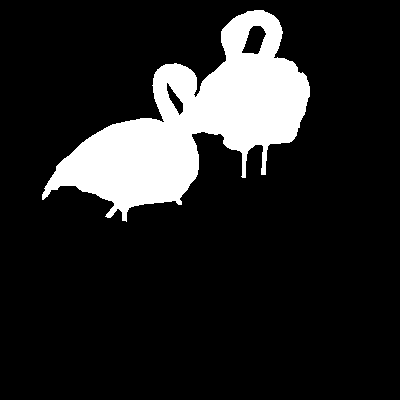}
        \textbf{GT}
    \end{minipage}
    \begin{minipage}[t]{0.1\textwidth}
    \vspace{0pt}
        \centering
        \includegraphics[width=1\linewidth]{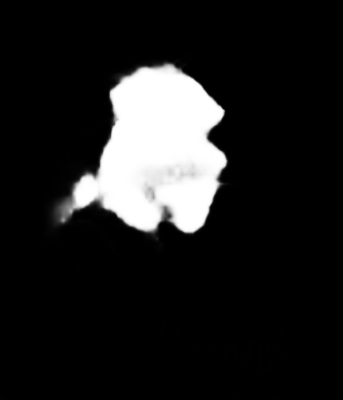}
        \includegraphics[width=1\linewidth]{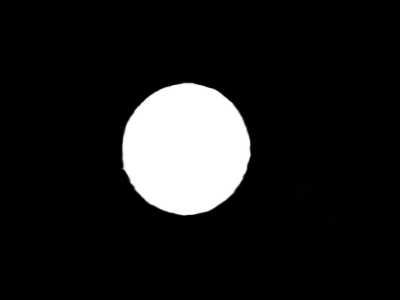}
        \includegraphics[width=1\linewidth]{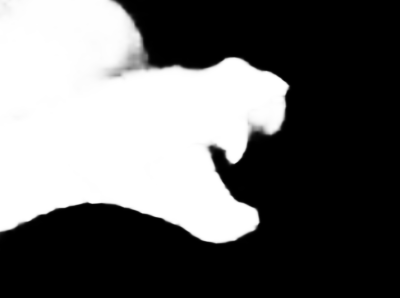}
        \includegraphics[width=1\linewidth]{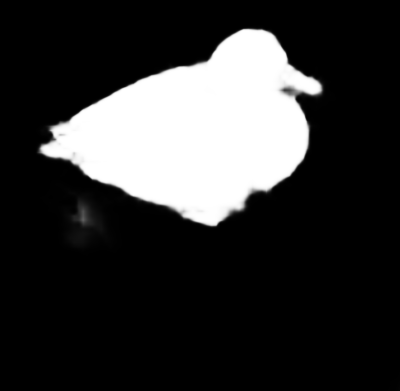}
        \includegraphics[width=1\linewidth]{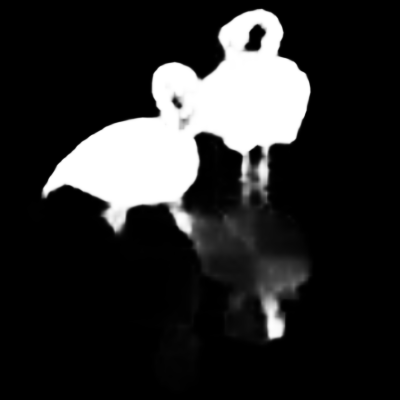}
        \textbf{Ours}
    \end{minipage}
    \begin{minipage}[t]{0.1\textwidth}
    \vspace{0pt}
        \centering
        \includegraphics[width=1\linewidth]{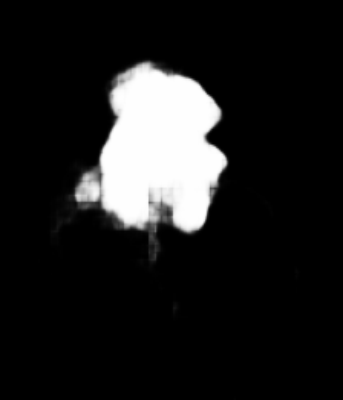}
        \includegraphics[width=1\linewidth]{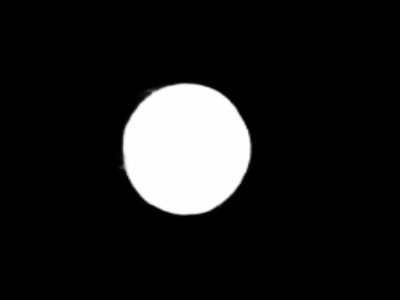}
        \includegraphics[width=1\linewidth]{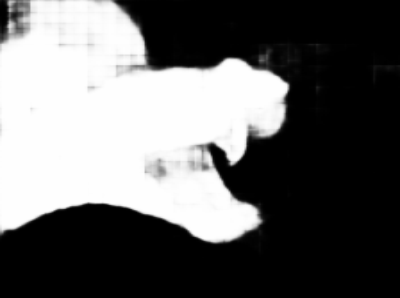}
        \includegraphics[width=1\linewidth]{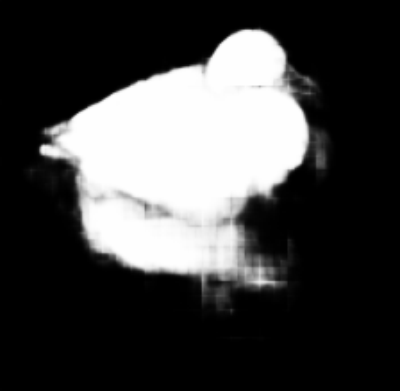}
        \includegraphics[width=1\linewidth]{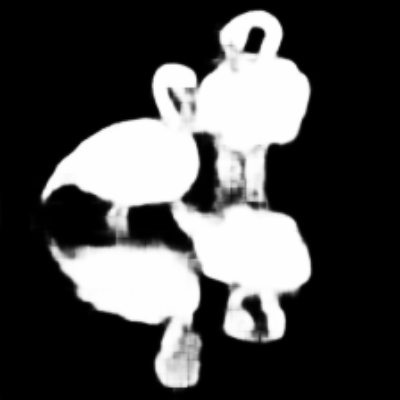}
        {VST}\cite{vst}
    \end{minipage}
        \begin{minipage}[t]{0.1\textwidth}
        \vspace{0pt}
        \centering
        \includegraphics[width=1\linewidth]{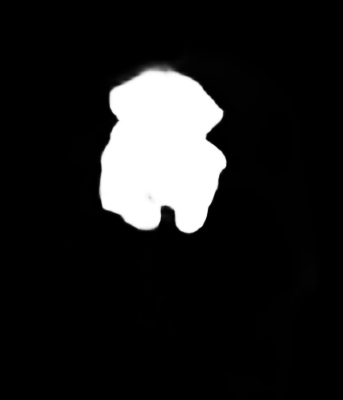}
        \includegraphics[width=1\linewidth]{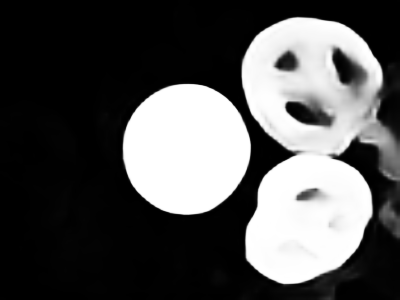}
        \includegraphics[width=1\linewidth]{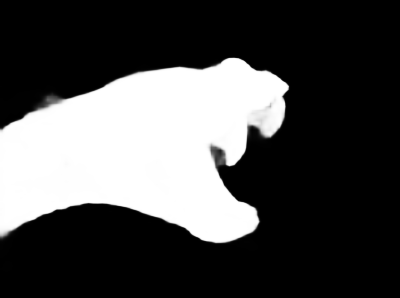}
        \includegraphics[width=1\linewidth]{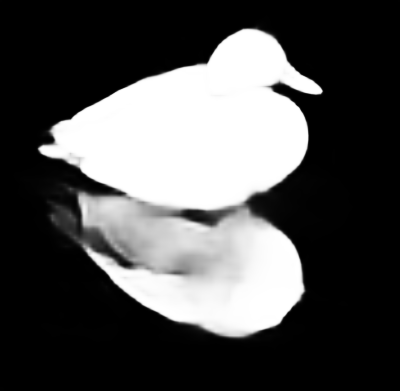}
        \includegraphics[width=1\linewidth]{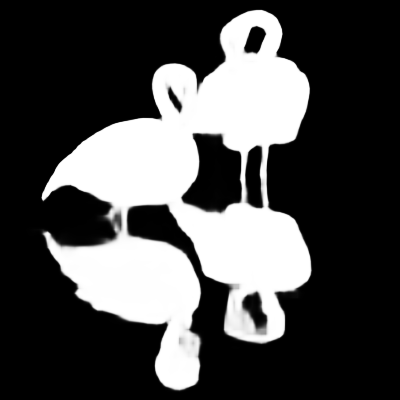}
        DAS\cite{dasnet}
    \end{minipage}
    \begin{minipage}[t]{0.1\textwidth}
    \vspace{0pt}
        \centering
        \includegraphics[width=1\linewidth]{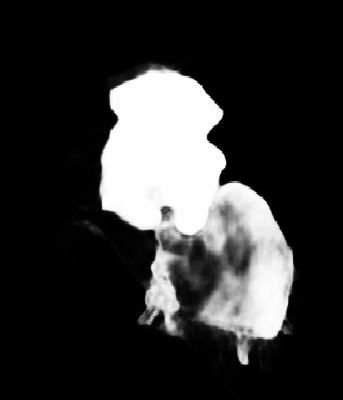}
        \includegraphics[width=1\linewidth]{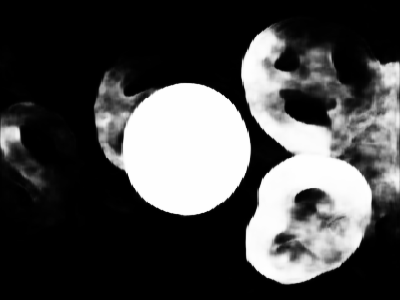}
        \includegraphics[width=1\linewidth]{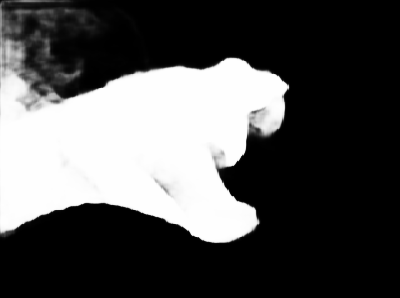}
        \includegraphics[width=1\linewidth]{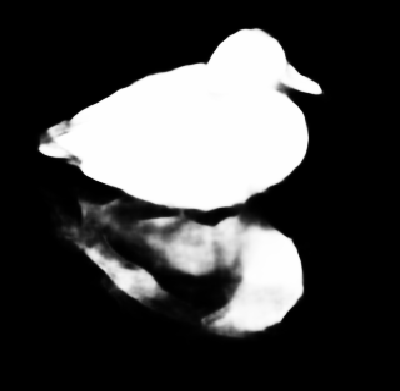}
        \includegraphics[width=1\linewidth]{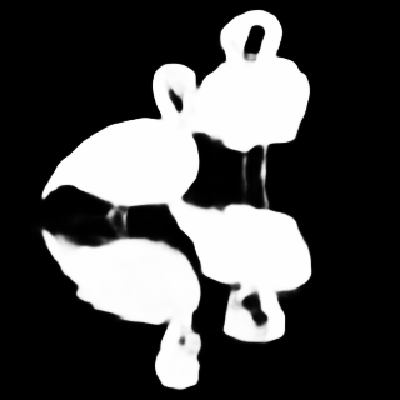}
        MINet\cite{MINet-R}
    \end{minipage}
    \begin{minipage}[t]{0.1\textwidth}
    \vspace{0pt}
        \centering
        \includegraphics[width=1\linewidth]{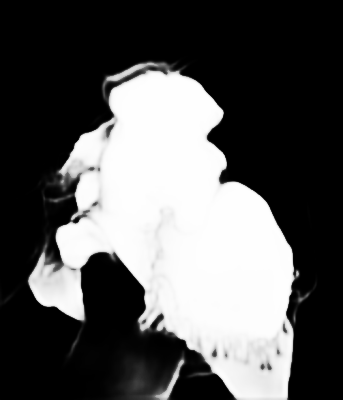}
        \includegraphics[width=1\linewidth]{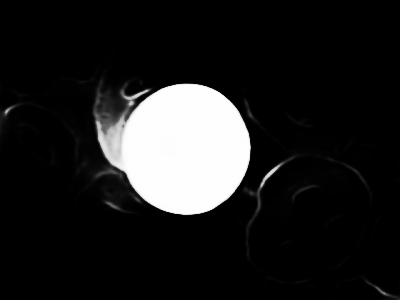}
        \includegraphics[width=1\linewidth]{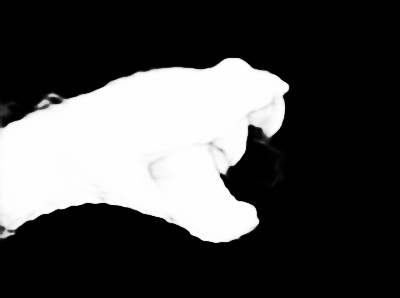}
        \includegraphics[width=1\linewidth]{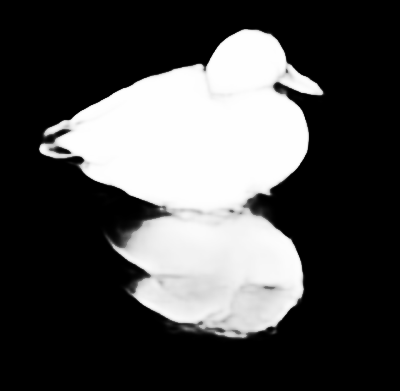}
        \includegraphics[width=1\linewidth]{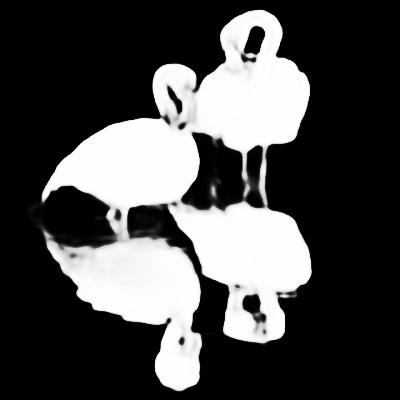}
        ITSD\cite{ITSD-R}
    \end{minipage}
    \begin{minipage}[t]{0.1\textwidth}
    \vspace{0pt}
        \centering
        \includegraphics[width=1\linewidth]{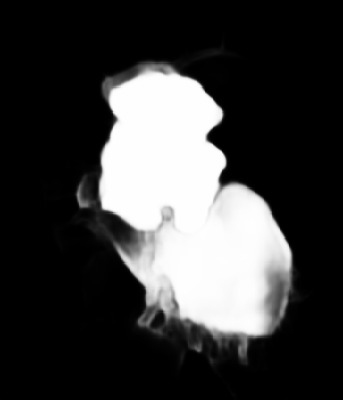}
        \includegraphics[width=1\linewidth]{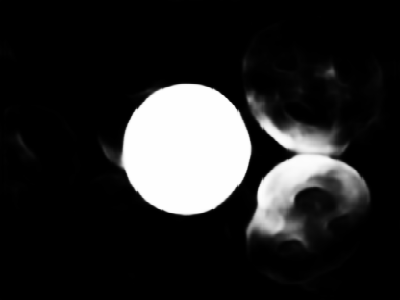}
        \includegraphics[width=1\linewidth]{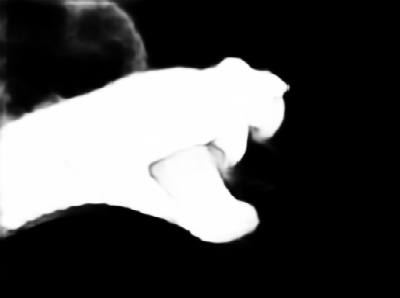}
        \includegraphics[width=1\linewidth]{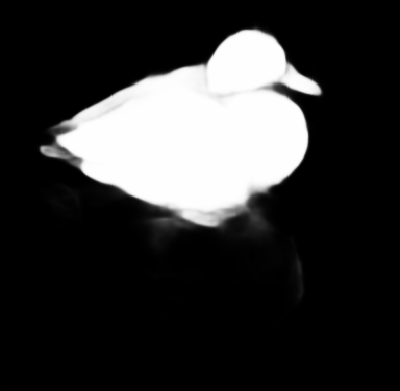}
        \includegraphics[width=1\linewidth]{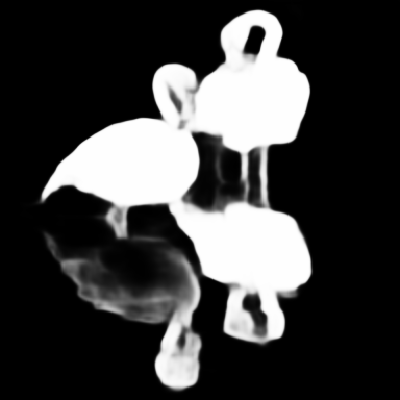}
        Gate\cite{gatenet}
    \end{minipage}
    \begin{minipage}[t]{0.1\textwidth}
    \vspace{0pt}
        \centering
        \includegraphics[width=1\linewidth]{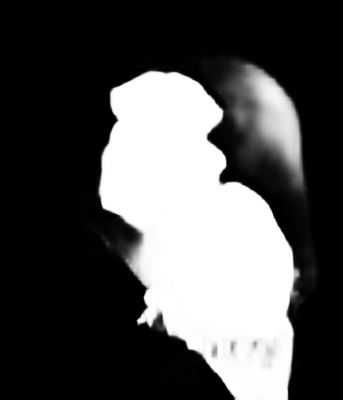}
        \includegraphics[width=1\linewidth]{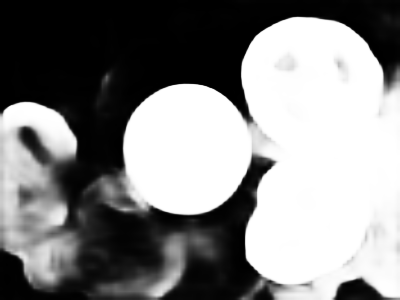}
        \includegraphics[width=1\linewidth]{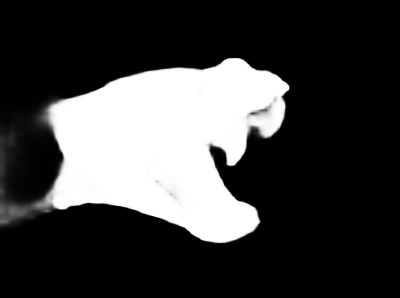}
        \includegraphics[width=1\linewidth]{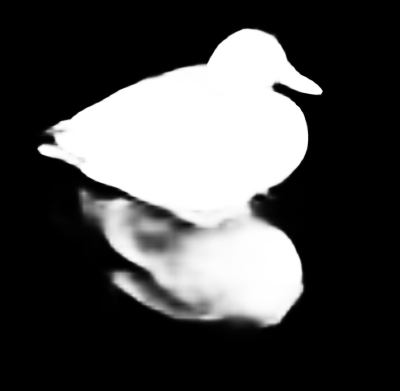}
        \includegraphics[width=1\linewidth]{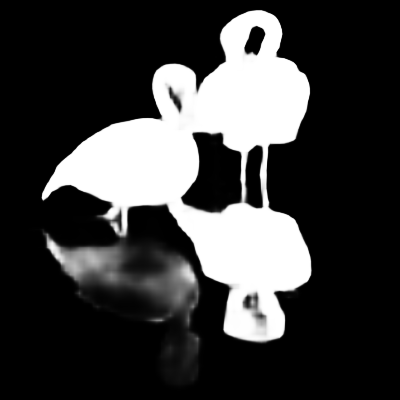}
        LDF\cite{LDF-R}
    \end{minipage}
    \caption{Qualitative comparison of our proposed method against the state-of-the-art \textbf{RGB} SOD methods.}
    \label{fig:visual_rgb}
\end{figure*}

Eight state-of-the-art methods, \emph{i.e.}, ITSD-R~\cite{ITSD-R}, MINet-R~\cite{MINet-R}, LDF-R~\cite{LDF-R}, CSF-R2~\cite{CSF-R2}, GateNet-R~\cite{gatenet}, DASNet~\cite{dasnet}, DH~\cite{vst}, and VST~\cite{vst}, are included for comparison in this experiment. The evaluation results are shown in Table~\ref{tab:rgb}. Similar trend of performance improvement to RGB-D-based comparison is observed. Particularly, our DFTR surpasses the state-of-the-art methods on most metrics (except $F_{\beta}$ on DUT-OMRON and HKU-IS, and $S_{\alpha}$ on PASCAL-S). We further visualize the saliency maps predicted by our model and other methods for qualitative comparison. As illustrated in Fig.~\ref{fig:visual_rgb}, our DFTR predicts more accurate saliency maps, validating the superiority of our model to other methods for salient object detection.

\begin{table*}[t]
    \centering{
    \renewcommand{\arraystretch}{1.05} 
    \renewcommand{\tabcolsep}{1mm} 
	\caption{Quantitative comparison of our DFTR with 8 state-of-the-art \textbf{RGB SOD} methods on 5 public RGB saliency benchmark datasets under 4 measurements (\emph{i.e.}, structure measure $S_{\alpha}$ \cite{fan2017structure}, maximum F-measure $F_{\beta}$ \cite{achanta2009frequency}, maximum enhanced alignment measure $E_{\xi}$\cite{Fan2018Enhanced}, and mean absolute error $M$ \cite{perazzi2012saliency}). ``$\uparrow$'' (or ``$\downarrow$'') indicates that larger (or smaller) is better. The subscript of each model denotes the publication year. In each column, the best result is marked with \red{red} and the second best with \blue{blue}.}
	\label{tab:rgb}
	\resizebox{\textwidth}{!}{
		\begin{tabular}{l|cccc|cccc|cccc|cccc|cccc}
			\hline\toprule
			\multicolumn{1}{c|}{\multirow{2}{*}{\textbf{Model}}} &
			\multicolumn{4}{c|}{ECSSD~\cite{ECSSD}} &
			\multicolumn{4}{c|}{DUTS~\cite{DUTS}} &
			\multicolumn{4}{c|}{DUT-OMRON~\cite{DUTOMRON}} &
			\multicolumn{4}{c|}{HKU-IS~\cite{HKUIS}} &
			\multicolumn{4}{c}{PASCAL-S~\cite{PASCAL}} \\ 
			&$S_{\alpha}\uparrow$   &$F_{\beta}\uparrow$    &$E_{\xi}\uparrow$  &$M\downarrow$
			&$S_{\alpha}\uparrow$   &$F_{\beta}\uparrow$    &$E_{\xi}\uparrow$  &$M\downarrow$
			&$S_{\alpha}\uparrow$   &$F_{\beta}\uparrow$    &$E_{\xi}\uparrow$  &$M\downarrow$
			&$S_{\alpha}\uparrow$   &$F_{\beta}\uparrow$    &$E_{\xi}\uparrow$  &$M\downarrow$
			&$S_{\alpha}\uparrow$   &$F_{\beta}\uparrow$    &$E_{\xi}\uparrow$  &$M\downarrow$  \\
			
			\midrule

			ITSD-R$_{20}$~\cite{ITSD-R}
			& .925 & .938 & .957 & .034
			& .885 & .867 & .929 & .041
			& .840 & .792 & .880 & .061
			& .917 & .926 & .960 & .031
			& .861 & .839 & .889 & .071\\

		    MINet-R$_{20}$~\cite{MINet-R}
		    & .925 & .938 & .957 & .034
		    & .884 & .864 & .926 & .037
		    & .833 & .769 & .869 & .056
		    & .919 & .926 & .960 & .029
		    & .856 & .831 & .883 & .071\\

		    LDF-R$_{20}$~\cite{LDF-R}
		    & .925 & .938 & .954 & {.034}
		    & .892 & .877 & .930 & .034
		    & .839 & .782 & .870 & .052
		    & .920 & .929 & .958 & .028
		    & .861 & .839 & .888 & .067\\
		    
		    CSF-R2$_{20}$~\cite{CSF-R2}
		    & .931 & .942 & .960 & .033
		    & .890 & .869 & .929 & .037
		    & .838 & .775 & .869 & .055
		    & - & - & - & -
		    & .863 &.839 & .885 & .073\\
		    
		    GateNet-R$_{20}$~\cite{gatenet}
		    & .924 & .935 & .955 & .038
		    & .891 & .874 & .932 & .038
		    & .840 & .782 & .878 & .055
		    & .921 & .926 & .959 & .031
		    & .863 & .836 & .886 & .071\\
		    
		    DASNet$_{20}$~\cite{dasnet}
			& .927 & .950 & - & \blue{.032}
			& .894 & {.896} & - & \blue{.034}
			& .845 & .827 & - & .050
			& .922 & .942 & - & .027
			& \red{.885} & .849 & - & \blue{.064}\\

			DH$_{21}$~\cite{DHNet}   
			& - & - & - & -
			& .892 & \red{.900} & - & .035
			& .843 & \red{.820} & - & \blue{.048}
			& .922 & \red{.944} & - & \blue{.026}
			& - & - & - & -\\
			
			VST$_{21}$~\cite{vst}   
			& \blue{.932} & \blue{.944} & \blue{.964} & .034
			& \blue{.896} & .877 & \blue{.939} & .037
			& \blue{.850} & .800 & \blue{.888} & .058
			& \blue{.928} & .937 & \blue{.968} & .030
			& .873 & \blue{.850} & \blue{.900} & .067\\	
			
			\midrule

			\textbf{DFTR~(Ours)}   
			& \red{.935} & \red{.949} &\red{.966} & \red{.028}
			& \red{.909} & \red{.900} & \red{.952} & \red{.029}
			& \red{.860} & \blue{.812} & \red{.895} & \red{.046}
			& \red{.930} & \blue{.941} & \red{.970} & \red{.025}
			& \blue{.879} & \red{.878} & \red{.924} & \red{.054}\\			
		
			\bottomrule\hline
	\end{tabular}}
	}
\end{table*}

\subsection{Ablation Study}\label{sec:ablation}

\begin{table}[t]
  \centering
    \renewcommand{\arraystretch}{1.05} 
    \renewcommand{\tabcolsep}{1mm} 
  \caption{Ablation study of different components of our DFTR. Model (a) denotes the baseline model with two linear layers followed by GELU activation as the saliency decoder. ``DS'' represents depth-supervision as an auxiliary task. ``MLS'' represents multi-level supervision. ``MFA'' and ``MFF'' are our proposed modules. We take \textbf{model (e)} as our final model. The best results are marked in \red{red}.}
    \label{tab:ab_model}
\resizebox{\textwidth}{!}{
    \begin{tabular}{c|cccc|cccc|cccc|cccc|cccc}
 \hline \toprule   
 \multicolumn{1}{c|}{\multirow{2}{*}{\textbf{Index}}} &
 \multicolumn{4}{c|}{\textbf{Combination}} & \multicolumn{4}{c|}{\textbf{NJU2K} \cite{fan2019rethinking}} &\multicolumn{4}{c|}{\textbf{NLPR} \cite{peng2014rgbd}}&\multicolumn{4}{c|}{\textbf{STERE}~\cite{niu2012leveraging}}&\multicolumn{4}{c}{\textbf{SIP}~\cite{fan2019rethinking}} \\
        & MFA & DS & MFF & MLS       
        &$S_{\alpha}\uparrow$   &$F_{\beta}\uparrow$    &$E_{\xi}\uparrow$  &$M\downarrow$
        &$S_{\alpha}\uparrow$   &$F_{\beta}\uparrow$    &$E_{\xi}\uparrow$  &$M\downarrow$
		&$S_{\alpha}\uparrow$   &$F_{\beta}\uparrow$    &$E_{\xi}\uparrow$  &$M\downarrow$
		&$S_{\alpha}\uparrow$   &$F_{\beta}\uparrow$    &$E_{\xi}\uparrow$  &$M\downarrow$
		\\
\midrule
    (a)     & & & & 
          & .869 & .863 & .926 & .066
          & .883 & .848 & .947 & .040
          & .878 & .864 & .932 & .059 
          & .826 & .814 & .902 & .081 
          \\
    (b)   & \checkmark & & &  
          & .916 & .919 & .951 & .937
          & .935 & .931 & .968 & .022
		  & .905 & .901 & .936 & .043
		  & .885 & .892 & .932 & .048
		  \\
          
    (c)    & \checkmark & \checkmark & &   
          & .919 & .923 & .953 & .036
          & .937 & .932 & .971 & .020
		  & .912 & .905 & .944 & .038
		  & .893 & .901 & .938 & .045
		  \\
          
    (d)    & \checkmark & \checkmark & \checkmark & 
          & .919 & .922 & .953 & .035
          & .937 & .933 & .970 & .020
		  & .915 & .910 & .949 & .035
		  & .902 & .910 & \red{.946} & .041
		  \\

    (e)    & \checkmark & \checkmark & \checkmark & \checkmark 
          & \red{.922} & \red{.923} & \red{.954} & \red{.034}
          & \red{.941} & \red{.934} & \red{.972} & \red{.018}
		  & \red{.918}& \red{.914} & \red{.951}   & \red{.034}
		  & \red{.904}& \red{.913} &\red{.946} &\red{.040}
		  \\

    \bottomrule     \hline
    \end{tabular}
    }
  \vspace{-0.25cm}
\end{table}

We conduct an ablation study to investigate the effectiveness of different components of the proposed DFTR on four commonly used datasets. It is worthwhile to mention that models under different settings are trained with the same protocol stated in Section.~\ref{sec:impl}. The evaluation results are presented in Table~\ref{tab:ab_model} and Table~\ref{tab:ab_para}.

\subsubsection{MFA Module.}
As shown in Table~\ref{tab:ab_model}, model (a), using Swin Transformer and MLP as the encoder and decoder, respectively, is adopted as the baseline. Here, (a) is an RGB-based SOD model. By switching the decoder from MLP to our multi-scale feature aggregation (MFA) module (\emph{i.e.}, model (b)), the SOD accuracies on all test sets are observed to significantly increase, which demonstrates the effectiveness of aggregating the information of multi-scale features for SOD.

\subsubsection{Depth Supervision.}
As previously mentioned, model (b) is an RGB-based framework. To evaluate the contribution of depth-supervised learning, we implement a multi-task framework (model (c)) by integrating the depth map prediction branch into model (b). The decoder for depth map prediction is the same as the SOD branch.
Due to the extra information provided by the depth map prediction task, the SOD performance of model (c) consistently surpasses that of model (b) as shown in Table~\ref{tab:ab_model}.

\subsubsection{MFF Module.}
To better fuse the features extracted from different streams, we propose a multi-stage feature fusion (MFF) module.
Based on model (c), we integrate MFF modules between the upper and lower MFA modules, which forms model (d). Such a module intends to facilitate the information flow between different branches. As Table~\ref{tab:ab_model} shows, model (d) only consistently surpasses model (c) on SIP~\cite{fan2019rethinking} while achieving comparable results on the other three datasets, It may be due to the fact that MFF module has a symmetric structure and thus tends to learn similar and redundant features. This issue, fortunately, can be mitigated by making use of the MLS.

\subsubsection{Multi-level Supervision (MLS).}
To validate the effectiveness of the one-side multi-level supervision (MLS) strategy adopted to MFF modules, we construct our final model (e) by adding MLS to model (d), as shown in Fig.~\ref{fig_network}. 
It can be observed from Table~\ref{tab:ab_model} that model (e) yields the best results under all metrics, which demonstrates that MLS does improve the final SOD performance by enforcing the upper output stream of the MFF module to focus more on salient regions at different network learning stages.
In summary, the above quantitative analysis shows the effectiveness of different components of our model. Our model is capable of integrating and fusing dual-stream multi-scale features and makes accurate dense SOD predictions.

\begin{table}[t]
  \centering
  \caption{Quantitative comparison with different hyper-parameter settings of our proposed DFTR model. Here, ``depth'' represents the depth of each Swin Transformer block in the MFA and MFF module, while ``down'' represents the ratio of input dimension and output dimension of linear layers in Eq.~\ref{formula:in_linear}. We take \textbf{setting (b)} as our final setting. \red{Red} denotes the best results.} 
  \label{tab:ab_para}
\setlength{\tabcolsep}{3pt} 
\renewcommand{\arraystretch}{1} 
   \resizebox{!}{1 cm}{
    \begin{tabular}{c|cc|cccc|cccc|cccc|cccc}
 \hline \toprule   
 \multicolumn{1}{c|}{\multirow{2}{*}{\textbf{Index}}} & 
 \multicolumn{2}{c|}{\textbf{Setting}} &  \multicolumn{4}{c|}{\textbf{NJU2K} \cite{fan2019rethinking}} &\multicolumn{4}{c|}{\textbf{NLPR} \cite{peng2014rgbd}}&\multicolumn{4}{c|}{\textbf{STERE}~\cite{niu2012leveraging}}&\multicolumn{4}{c}{\textbf{SIP}~\cite{fan2019rethinking}} \\
        &  depth & down      
        &$S_{\alpha}\uparrow$   &$F_{\beta}\uparrow$    &$E_{\xi}\uparrow$  &$M\downarrow$
        &$S_{\alpha}\uparrow$   &$F_{\beta}\uparrow$    &$E_{\xi}\uparrow$  &$M\downarrow$
		&$S_{\alpha}\uparrow$   &$F_{\beta}\uparrow$    &$E_{\xi}\uparrow$  &$M\downarrow$
        &$S_{\alpha}\uparrow$   &$F_{\beta}\uparrow$    &$E_{\xi}\uparrow$  &$M\downarrow$		\\
\midrule 
    (a)   & 2 & 8
          & .918 & .920 & .951 & .038
          & .938 & .933 & .970 & {.019}
		  & .916 & .912 & .948 & .035
		  & .902 & .911 & .947 & {.042}
          \\
                    
    (b)   & 1 & 8
          & \red{.922} & .923 & .954 & \red{.034}
          & \red{.941} & \red{.934} & \red{.972} & \red{.018}
		  & \red{.918} & \red{.914} & \red{.951} & \red{.034}
		  & \red{.904} & \red{.913} & .946 & \red{.040}
		  \\
          
    (c)   & 1 & 4
          & .918 & .920 & .952 & {.037}
          & .937 & .930 & .967 & .021
		  & .914 & .909 & .942 & .039
		  & .899 & .908 & .940 & .046
		  \\
          
    (d)   & 1 & 16
          & .921 & \red{.924} & \red{.955} & \red{.034}
          & .939 & .933 & .971 & \red{.018}
		  & .917 & .912 & .949 & .035
		  & .903 & .912 & \red{.948} & .041
		  
		  \\

    \bottomrule     \hline
    \end{tabular}
    }
  \vspace{-0.25cm}
\end{table} 

\subsubsection{Evaluation on Hyper-parameters of DFTR}
Apart from different modules and strategies adopted in our DFTR, the setting of hyper-parameters is also an important factor, which may affect the model performance. To this end, we conduct an experiment to evaluate the model performance with different hyper-parameter settings.

The ``depth'' and ``down'' are two main hyper-parameters of our DFTR. Concretely, ``depth'' $\in \{1,2\}$ represents the depth of Swin Transformer block adopted by MFA and MFF modules and 
``down'' $\in \{4,8,16\}$ represents the ratio of input dimension and output dimension of linear layers in Eq.~(\ref{formula:in_linear}). The evaluation results are presented in Table~\ref{tab:ab_para}. As shown, setting (b), \emph{i.e.}, depth=1 and down=8, reaches the best results with most metrics, which indicates the effectiveness of the linear layer for channel dimensional reduction.
Besides, the 1-depth Swin Transformer block is able to learn local features without shifting windows. 

\section{Conclusion}
\label{sec:concl}
We reclaimed the superiority of depth-supervised SOD framework through the utilization of Transformer architecture, and developed a pure Transformer-based network DFTR that works for both RGB and RGB-D SOD tasks.
The decoder of DFTR consists of an MFA module that gradually aggregates adjacent-scale features from coarse to fine and an MFF module that bidirectionally fuses the decoding features in a hierarchical manner with the emphasis on salient regions. Quantitative and qualitative results on both RGB and RGB-D SOD benchmarks demonstrated the effectiveness of our DFTR. With only RGB inputs at the inference stage, our DFTR detects salient objects more accurately than any other CNN-based method or Transformer-based method.
\clearpage
%
%
\bibliographystyle{splncs04}
\bibliography{paper}
\end{document}